\documentclass{article}
\pdfoutput=1
\usepackage{PRIMEarxiv}

\usepackage[utf8]{inputenc} 
\usepackage[T1]{fontenc}    
\usepackage{url}            
\usepackage{booktabs}       
\usepackage{amsfonts}       
\usepackage{nicefrac}       
\usepackage{microtype}      
\usepackage{lipsum}
\usepackage{fancyhdr}       
\usepackage{graphicx}       

\usepackage{cite}
\usepackage{babel}
\usepackage{amsmath,amssymb,amsfonts}
\usepackage{algorithmic}
\usepackage[]{algorithm2e}
\usepackage{graphicx}
\usepackage{textcomp}
\usepackage{graphicx,wrapfig}
\usepackage{subcaption}
\usepackage{mwe}
\usepackage{flushend}
\usepackage{tikz, pgfplots, pgfplotstable, schemabloc}
\usetikzlibrary{shapes,shapes.geometric,arrows,fit,calc,positioning,automata,patterns}
\usepackage{quoting,xparse}
\usepackage{listings}
\usepackage{setspace}
\usepackage{tabularx}
\usepackage{booktabs}
\usepackage{longtable}
\usepackage{multirow}
\usepackage{array}
\usepackage{hyperref}
\usepackage{blindtext}
\usepackage{threeparttable, tablefootnote}
\usepackage{endnotes}
\usepackage{marvosym}

\pgfplotsset{compat=1.17}
\DeclareUnicodeCharacter{2212}{\textendash}

\newcolumntype{L}[1]{>{\raggedright\arraybackslash}p{#1}}
\newcolumntype{C}[1]{>{\centering\arraybackslash}p{#1}}
\newcolumntype{R}[1]{>{\raggedleft\arraybackslash}p{#1}}
\definecolor{rosso}{RGB}{220,57,18}
\definecolor{giallo}{RGB}{255,153,0}
\definecolor{blu}{RGB}{102,140,217}
\definecolor{verde}{RGB}{16,150,24}
\definecolor{viola}{RGB}{153,0,153}
\definecolor{ggrigio}{RGB}{200,200,200}

\definecolor{bblue}{HTML}{4F81BD}
\definecolor{rred}{HTML}{C0504D}
\definecolor{ggreen}{HTML}{9BBB59}
\definecolor{ppurple}{HTML}{9F4C7C}

\definecolor{palette_blue}{HTML}{20639B}

\newcounter{groupcount}
\pgfplotsset{
    draw group line/.style n args={5}{
        after end axis/.append code={
            \setcounter{groupcount}{0}
            \pgfplotstableforeachcolumnelement{#1}\of\datatable\as\cell{%
                \def\temp{#2}
                \ifx\temp\cell
                    \ifnum\thegroupcount=0
                        \stepcounter{groupcount}
                        \pgfplotstablegetelem{\pgfplotstablerow}{X}\of\datatable
                        \coordinate [yshift=#4] (startgroup) at (axis cs:\pgfplotsretval,0);
                    \else
                        \pgfplotstablegetelem{\pgfplotstablerow}{X}\of\datatable
                        \coordinate [yshift=#4] (endgroup) at (axis cs:\pgfplotsretval,0);
                    \fi
                \else
                    \ifnum\thegroupcount=1
                        \setcounter{groupcount}{0}
                        \draw [
                            shorten >=-#5,
                            shorten <=-#5
                        ] (startgroup) -- node [anchor=base, yshift=0.5ex] {#3} (endgroup);
                    \fi
                \fi
            }
            \ifnum\thegroupcount=1
                        \setcounter{groupcount}{0}
                        \draw [
                            shorten >=-#5,
                            shorten <=-#5
                        ] (startgroup) -- node [anchor=base, yshift=0.5ex] {#3} (endgroup);
            \fi
        }
    }
}

\title{An adaptive cognitive sensor node for ECG monitoring in the Internet of Medical Things}

\author{
  Matteo A. Scrugli, Daniela Loi, Luigi Raffo, Paolo Meloni \\
  University of Cagliari \\
  \texttt{\{matteo.scrugli, daniela.loi, raffo, paolo.meloni\}@unica.it} \
}

\begin{document}
\maketitle

\begin{abstract}
\label{sec:abstract}
The Internet of Medical Things (IoMT) paradigm is becoming mainstream in multiple clinical trials and healthcare procedures. Cardiovascular diseases monitoring, usually involving electrocardiogram (ECG) traces analysis, is one of the most promising and high-impact applications. 
Nevertheless, to fully exploit the potential of IoMT in this domain, some steps forward are needed. First, the edge-computing paradigm must be added to the picture. A certain level of near-sensor processing has to be enabled, to improve the scalability, portability, reliability, responsiveness of the IoMT nodes. Second, novel, increasingly accurate, data analysis algorithms, such as those based on artificial intelligence and Deep Learning, must be exploited. To reach these objectives, designers and programmers of IoMT nodes, have to face challenging optimization tasks, in order to execute fairly complex computing tasks on low-power wearable and portable processing systems, with tight power and battery lifetime budgets.
In this work, we explore the implementation of a cognitive data analysis algorithm, based on a convolutional neural network trained to classify ECG waveforms, on a resource-constrained microcontroller-based computing platform. To minimize power consumption, we add an adaptivity layer that dynamically manages the hardware and software configuration of the device to adapt it at runtime to the required operating mode. Our experimental results show that adapting the node setup to the workload at runtime can save up to 50\% power consumption. Our optimized and quantized neural network reaches an accuracy value higher than 97\% for arrhythmia disorders detection on MIT-BIH Arrhythmia dataset.
\end{abstract}


\section{Introduction}
\label{sec:introduction}
The Internet-of-Things (IoT) paradigm, declined in the so-called Internet of Medical Things (IoMT), enables seamless collection of a wide range of data streams, that can be analyzed to extract relevant information about the patient's condition.
However, in order to make IoMT really ubiquitous and effective, a step forward is needed to improve scalability, responsiveness, security, privacy. Most of the efforts aiming in this direction focus on the adoption of an edge-computing approach. Data streams, acquired by sensors, can be processed, at least partially, at the edge, before being sent to the cloud, on adequate portable/wearable processing platform. This provides several advantages. First, it reduces bandwidth requirements. Near-sensor processing can extract from raw data more compact information. In this way, less communication bandwidth is required to the centralized server, and, at the same time, the energy consumption related to wireless data transmission is drastically reduced. Second, near-sensor processing can improve reliability. Monitoring must not rely necessarily on connection availability and, if immediate feedback to the user and/or local actuation is needed, the delays through the network can be avoided. Moreover, pre-processed information can be delivered to the cloud, preserving user privacy avoiding the propagation of sensitive raw data. \par
An extremely important field of application of IoMT is related to the treatment of cardiovascular diseases (CVD), a major public health problem that generates millions of deaths yearly and impacts significantly on health-related public costs. As an example, in 2016, $\approx17.6$ million (95\% CI, 17.3–18.1 million) deaths were attributed to 
CVD globally, representing an increase of 14.5\% (95\% CI, 12.1\%–17.1\%) since 2006\cite{cvd}. In Europe, the CVD impact on the economy is estimated to be around \texteuro 210 billion\cite{european}.
CVD treatment with remote monitoring involves in most cases analysis of electrocardiogram (ECG) signals. Creating embedded platforms implementing such kind of analysis is promising, but, at the same time, very challenging, for several reasons: 
\begin{itemize}
    \item \textbf{Requires edge computing at low energy/cost budget: } Sensor nodes must be wearable and affordable to implement ubiquitous patient monitoring. Given the high data rate produced by ECG sensors, raw data wireless data transmission requires an energy budget that cannot be negligible when the task is implemented in a portable and inexpensive computing device. 
    \item \textbf{Requires cognitive computing:} state-of-the-art anomalies detection tasks are based on the analysis of manually designed features with are hard to craft and extract online from the ECG waveforms. Thus the community is shifting focus to techniques based on neural networks and deep learning, that rely on automatically learned features. However, existing approaches that use deep learning for the recognition of anomalies on the ECG trace, rarely pay attention to energy consumption to be deployed on low-power processing systems. Thus, pretty often do not take into account workload reduction and post-deployment accuracy evaluation. 
    \item \textbf{Requires adaptivity:} Intensity of the processing workload is very dependent on the needed level of detail and also intrinsically data-dependent. Information to be analyzed is usually contained in waveform shapes of ECG peaks, thus the rate of sample frames to be analyzed is directly dependent on the patient's heartbeat rate. This paves the way to energy consumption reduction by means of an adaptive management of the system, that reconfigures itself on the basis of the detected data and on the chosen operating mode. 
\end{itemize}



In this work, we explore the implementation of a system for at-the-edge cognitive processing of ECG data. We have conceived
a hardware/software setup for the processing system inside the IoMT node. We have used SensorTile, a compact processing device developed by STMicroelectronics, as a reference microcontroller platform. 
The system makes use of a quantized convolutional neural network, specifically sized and trained to run on a low-power microcontroller, that has been validated in post-deployment and recovers accuracy drops that arise in real on-line utilization.
Moreover, we take a step further in hardware/software optimization using adaptivity, allowing the system to reconfigure itself, to suit different operating modes and data processing rates.
To this aim, besides executing the tasks that implement sensor monitoring and on-board processing, the system includes a component called ADAM (ADAptive runtime Manager), able to dynamically manage the hardware/software configuration of the device optimizing power consumption and performance. ADAM creates and manages a network of processes that communicate with each other via FIFOs. The morphology of the process network varies to match the needs of the operating mode in execution. ADAM can be triggered by re-configuration messages sent by the external environment or by specific workload-related variables in the sampled streams (e.g. patient's heartbeat pace). When triggered, ADAM changes the morphology of the process network, switching on or off processes, and reconfigures the inter-process FIFOs. Moreover, depending on the new configuration it changes the hardware setup of the processing platform, adapting power-relevant settings such as clock frequency, supply voltage, peripheral gating.


The remainder of this paper is as follows: Section \ref{sec:relatedwork} describes the landscape of related work in literature, Section \ref{sec:system} describes an overview of the overall SoS picture, Section \ref{sec:IoMT_node} present the proposed template for the node and the reference target platform and the reference application model. Moreover it presents the details of the ADAM component. Section \ref{sec:usecase} describes how the chosen template has been declined to implement ECG monitoring, the proposed operating modes and the processing tasks coexisting in the application. Section \ref{sec:results} discusses our experimental results. Finally, Section \ref{sec:conclusion} outlines our conclusions. 

\section{Related work}
\label{sec:relatedwork}
Multiple solutions involving the use of sensor networks in hospitals or at home and the IoMT are proposed in literature \cite{net1, net3, Macis15}. Most of these studies exploit a cloud-based analysis: data is usually encapsulated in standard formats and sent to remote servers for data mining. 
Most research work takes into account wearability and portability as main objectives when developing IoMT-based data sensing architectures, thus devices available on the market can guarantee autonomy for days or weeks \cite{fit1, fit2}.\par
To really use cognitive computing at the edge, more complex and accurate algorithms, such as those exploiting artificial intelligence or deep learning, must be targeted. Their efficiency has been widely demonstrated on high-performance computing platforms. Some examples are \cite{heavy1}, where an NVIDIA GeForce GTX 1080 Ti (11 GB) is used, \cite{heavy2}, that uses a $3.5\,G\!H\!z$ Intel Core i7-7800X CPU, RAM 32 GB, and a GPU NVIDIA Titan X (Pascal, 12 GB), or \cite{heavy3}, based on an i7-4790 CPU at $3.60\,G\!H\!z$.
However, how to map state of the art cognitive computing on resource-constrained platforms
is still an open question. There is an ever-increasing number of approaches focusing on machine learning and artificial intelligence to identify specific events in sensed data. In \cite{nodata} and \cite{bologna} authors exploit ANN (artificial neural networks) to detect specific conditions from the proposed data. In \cite{bologna}, an ANN is used to identify the emotional states (happiness or sadness) of the patient. However, network topologies are still very basic and highly tuned and customized to fit on the target device. 
In \cite{prev}, energy/power efficiency is improved, using near-sensor processing to save data transfers, and dynamically adapting application setup and system frequency to the operating mode requested by an external user and to data-dependent workload. As a use-case, a CNN (Convolutional Neural Network) is used to identify anomalies on ECG traces.\par
There are several works that implement ECG monitoring on customized chips, it is shown that with low energy consumption it is possible to classify cardiac anomalies in real-time even using AI methods \cite{custom0, custom1, custom2, custom3, custom4, custom5, custom6}. Other work focuses on implementing efficient off-the-shelf commercial devices to facilitate easier community adoption of these techniques. Several target technologies have been used in the literature, such as FPGAs or microcontroller-based boards.\par
A substantial number of research works are dedicated to studying IoT devices in the medical field, in particular ECG monitoring and detection of anomalies \cite{other1,other2,other3,other4,other5,other6,other7,other8}.
The edge-computing paradigm is often only marginally exploited, and local processing is used only for implementing easy checks on raw data and/or marshaling tasks for wrapping the sensed data inside standard communication protocols \cite{lp1,lp2,lp3,lp4}.
To really use cognitive computing at the edge, more complex and accurate algorithms, such as those exploiting artificial intelligence or deep learning, must be targeted.\par
The cognitive approach that involves the use of convolutional neural networks (CNNs) shows promise in terms of accuracy in detecting ECG signal arrhythmias compared to other traditional strategies based or not on artificial intelligence algorithms \cite{ieee1,ieee2,ieee6}, Moreover, in most cases, the use of CNNs allows to classify an ECG signal even if not pre-processed.
The most common strategies present in many state-of-the-art works that allow to improve the efficiency of these IoT nodes are: moving the inference operations at the edge, choosing a low-power device, quantization techniques to speed up the network execution of the inference stage.\par
Another interesting work was presented in \cite{ieee2}, in addition to the comparison with other techniques used to analyze the ECG trace, Latent Semantic Analysis techniques were used to improve the accuracy of the network. Both training and inference take place on the cloud side, our aim is to move the inference to the edge of a low-power device in order to reduce latency times and reduce energy consumption due to wireless communication.\par
In \cite{acc} excellent results were obtained for ventricular arrhythmias and supraventricular arrhythmias classification: 99.6\% and 99.3\% for accuracy value, 98.4\% and 90.1\% for sensitivity value, 99.2\% and 94.7\% for positive predictive value, respectively. In \cite{acc} a double CNN is used, one of them takes as input the frequency domain information of the ECG signal (a fast Fourier transform is performed). This methodology, despite the excellent results in terms of accuracy, was not taken into consideration in our case because it's particularly expensive to perform on a microcontroller.\par
In \cite{ieee6}, again, there is proof of how neural networks obtain good results if compared with methods such as K-nearest neighbors (KNN) and random forest (RF) (95.98\% on MIT-BIH Supraventricular Arrhythmia Database) and the inference occurs directly from the IoT node but the power consumption remain relatively high once again, they are used in fact non-low power devices such as Raspberry Pi 4 or Jetson Nano. Always in \cite{ieee6}, a good job of research has been done on the morphology of the CNN network that was more suitable for inference on ECG signals, the network we used provides a structure very similar to the one chosen in \cite{ieee6}.\par
In \cite{ieee5}, good results are obtained in terms of accuracy (96\% using MIT Arrhythmia dataset), but here too the inference occurs on the cloud side, albeit with excellent latency times the node will still have to transmit a large amount of raw data which in our case causes excessive consumption of power with respect having edge-side inference.\par
An approach similar to \cite{ieee3} was chosen, an embedded device was chosen that is able to perform the inference directly on the node. In \cite{ieee3}, a study was made on the variation of accuracy as a function of different quantization levels, they choose a precision of 12-bit with an accuracy of 97\%, but it's visible that already from 6-bit upwards the accuracy levels exceed the 90\%. Power consumption is around $200\,mW$ during computation and the node is based on FPGA technology.\par
Other works with which we are confronted are \cite{rbm, san, dcnn, rasp}.
The quantization technique will also be exploited in our work by choosing an 8-bit precision, this will allow us to significantly speed up the inference operations. In particular, the CMSIS libraries are used to exploit the SIMD capacity of the microcontroller chosen.\par
In this work, we extend \cite{prev} taking into account that in the current state-of-the-art landscape, network topologies, processing platforms and software tools can be much more complex. On one hand, the community has designed novel ultra-low power processing platforms, providing previously unmatched computation capabilities on typical AI and data analysis workloads.
In summary, as main novel contribution, we propose:
\begin{itemize}
    \item The definition of a hardware/software/firmware architectural template for the implementation of a remotely-controlled sensory node, allowing for near-sensor cognitive data processing, inserted in an IoT context.
    \item Its validation on a state-of-the-art data analysis based on a Convolutional Neural Network as an example computational load.
    \item The evaluation of the effectiveness of in-place computing and operating mode dynamic optimization on ARM microcontroller platform, as a method to reduce the power consumption of the node, on a case study involving classification of ECG data.
\end{itemize}

\begin{table*}
	\centering
	\scriptsize
    \begin{longtable}{C{10mm} C{38mm} C{23mm} C{13mm} C{38mm} C{15mm}}
		\hline
		\\[-1.6mm]
		\textit{\textbf{Reference}} &
		\textit{\textbf{Node technology}} &
		\textit{\textbf{Processing placement}} &
		\textit{\textbf{Power consumption}} &
		\textit{\textbf{Accuracy}} &
		\textit{\textbf{Classification method}} \\
		\\[-2mm]
		\hline
		\hline
		\\[0mm]
		
		
		\textit{\textbf{\cite{acc}}} &
		$ - $ &
		$ - $ &
		$ - $ &
		MIT-BIH ventricular arrhythmias and supraventricular arrhythmias classification: 99.6\% and 99.3\% for accuracy value, 98.4\% and 90.1\% for sensitivity value, 99.2\% and 94.7\% for precision value, respectively &
		CNN \\
		\\[-0mm]
		
		\textit{\textbf{\cite{ieee2}}} &
		$ - $ &
		training and inference on cloud  &
		$ - $ &
		94\% for sensitivity vale, 99,3\% for accuracy value on custom dataset & CNN + LSA method \\
		\\[-0mm]
		
		\textit{\textbf{\cite{rbm}}} & $-$ & $-$ & $-$ & accuracies of 93.63\% for ventricular ectopic beats and  95.57\% for supraventricular ectopic beats on MIT Arrhythmia dataset & RBM and DBN \\
		\\[-0mm]
		
		\textit{\textbf{\cite{san}}} & $-$ & $-$ & $-$ & MIT-BIH dataset, NSVFQ classes: 99,09\%, 98,55\% and 99,52\% for accuracy, sensitivity and specificity value, respectively & DNN \\
		\\[-0mm]
		
		\textit{\textbf{\cite{dcnn}}} & Intel I7-4700MQ at 2.4 GHz (eight CPUs) and 16-Gb memory, but designed to run on cheaper and less powerful architectures & inference on edge & $-$ & accuracy up to 99\% for ventricular ectopic beats and up to 97.6\% for supraventricular ectopic beats on MIT Arrhythmia dataset & adaptive implementation of 1-D CNNs\\
		\\[-0mm]
		
		\textit{\textbf{\cite{ieee5}}} &
		hierarchical structure &
		training on cloud, inference on edge &
		$ - $ &
		96\% for accuracy value on MIT Arrhythmia dataset & CNN \\
		\\[0mm]
		
		\textit{\textbf{\cite{ieee3}}}  &
		FPGA &
		inference on edge &
		$ 200\,mW $ $^\text{I}$ &
		97\% accuracy value for NLRAV classes on MIT Arrhythmia dataset &
		quantized CNN \\
		
		\\[0mm]
		
		\textit{\textbf{\cite{ieee6}}} &
		Jetson Nano (Quad-core ARM A57 @ 1.43GHz), Raspberry Pi 4 (Quad-core CortexA72 @ 1.5GHz), Raspberry Pi 3 (Quad-core Cortex-A53 @ 1.4GHz) & inference on edge &
		$ - $ &
		95.27\% accuracy value for NSVF classes on MIT-BIH Supraventricular Arrhythmia Database & CNN \\
		
		\\[0mm]
		
		\textit{\textbf{\cite{rasp}}} &
		Raspberry Pi 3 &
		training on cloud, inference on edge &
		$ - $ &
		96\% for accuracy value for: normal(NOR), Left Bundle Brunch Block(LBB), Right Bundle Brunch Block (RBB), Paced beat(PAB), Premature Ventricular Contraction(PVC), Atrial Premature Contraction(APC), Ventricular Flutter Wave(VFW) and Ventricular Escape Beat(VEB) beats for detecting arrhythmia disorders on MIT Arrhythmia dataset & CNN \\
		
		\\[0mm]
		
		\textit{\textbf{Our work}} &
		ST SensorTile &
		inference on edge &
		$ 9\,mW $ $^\text{II}$  &
		MIT-BIH dataset, NLRAV and NSVFQ classes: 97.42\% and 96.98\% for accuracy value, 98.26\% and 98.22\% for sensitivity value, 98.28\% and 98.52\% for precision value, respectively &
		quantized CNN \\
		
        \\[0mm]
		\hline
		\\[0mm]
		\multicolumn{6}{l}{$^\text{I}$When fully active.}\\
		\multicolumn{6}{l}{$^\text{II}$The device adapts itself to the workload and operating mode, the reported value is the power consumption in case of maximum workload.}\\
		\\[0mm]
		\hline
		\\[0mm]
    \end{longtable}
	\caption{Structure of some devices proposed in literature.}
	\label{tab:literature}
\end{table*}


\section{Adaptive sensor node architecture}
\label{sec:system}
\begin{figure}
    \centering
    \begin{minipage}{0.47\textwidth}
        \centering
        \includegraphics[width=1\textwidth]{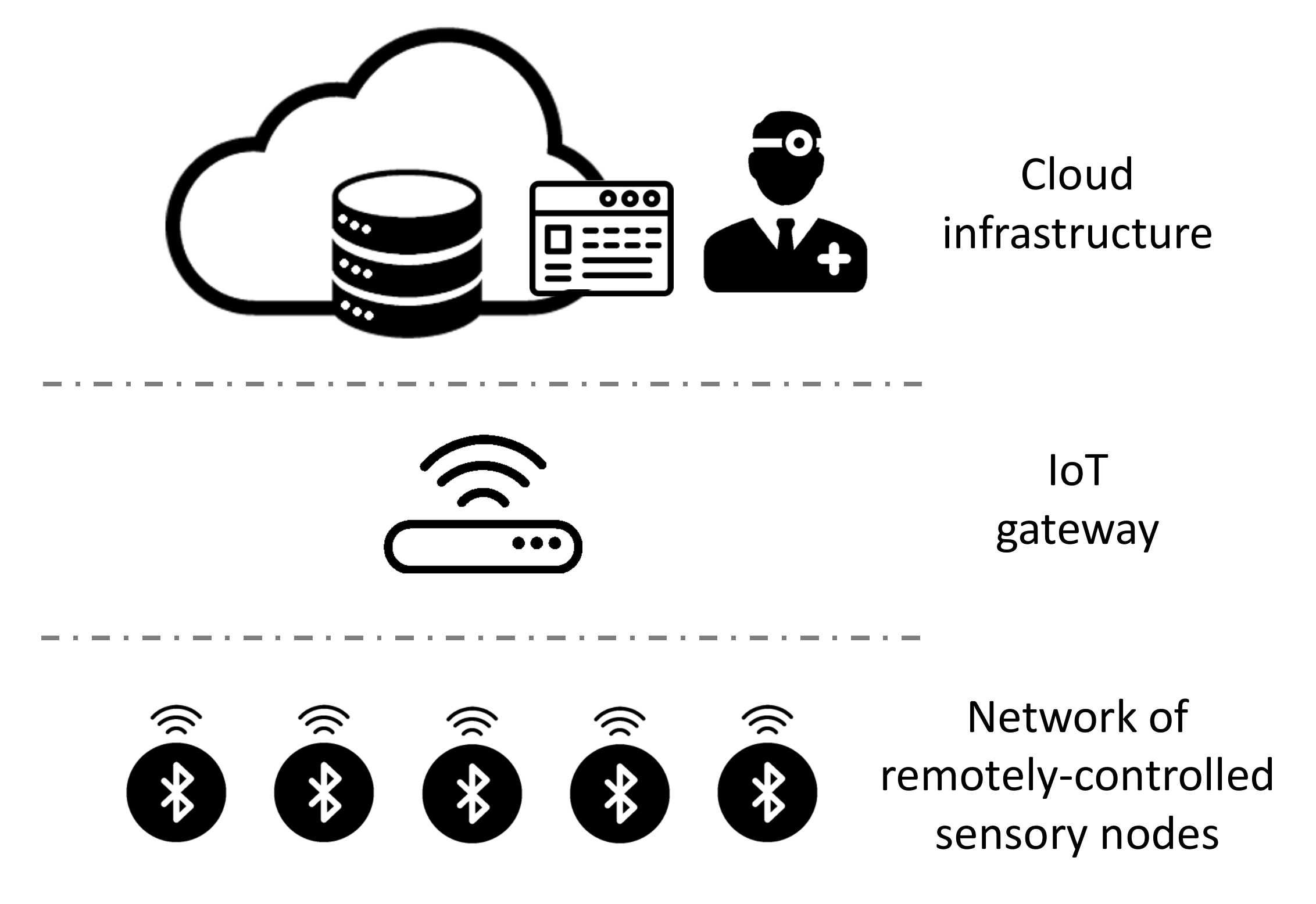}
  \caption{General overview of the proposed system.}
  \label{fig:INSIEME}
    \end{minipage}
\end{figure}

Figure \ref{fig:INSIEME} shows an overview of the system architecture as envisioned in this work.
We see the network as composed of three levels. The lower level is composed of the sensor nodes, which acquire information from the environment. They are connected to the upper level using  Bluetooth technology. The nodes are capable of reacting, reconfiguring their operating mode, to commands sent from higher levels, or to workload changes that can be detected near-sensor, thanks to the internal component called ADAptive runtime Manager which will be described later.
The intermediate level consists of several gateways, in charge of collecting the data from the sensor nodes and send them to the upper level. To test the approach presented in this work, the gateway was implemented with a Raspberry Pi 3 running a Linux operating system.

For the same purpose, the cloud-based infrastructure, on top of the stack, has been implemented using Google App Engine. Data is stored securely on the cloud, and can be used for analysis or simply visualized by a healthcare professional. Such kind of user, accessing a web-based interface, can also send downstream commands to the nodes, to communicate a required change of the operating mode, e.g. changing the needed detail of acquisition of the patient's parameters.

In this paper, attention will be focused only on the sensor node.

\section{IoMT Node architecture}
\label{sec:IoMT_node}
\begin{figure}
  \centering
  \includegraphics[width=0.4\textwidth]{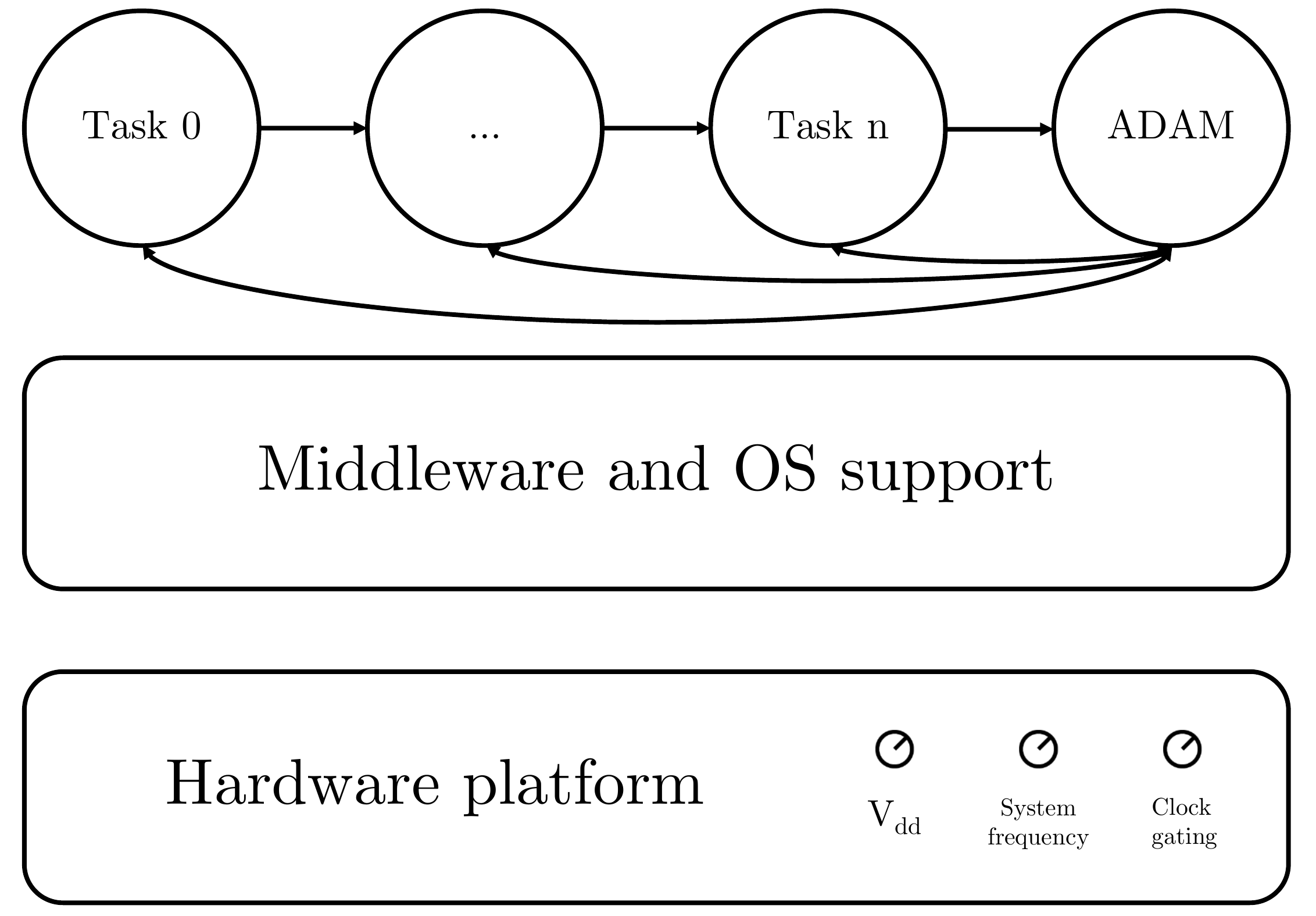}
          \caption{IoMT node architecture overview.}
   \label{fig:node}
\end{figure}

The sensor node architecture itself can be seen as a layered structure, schematized in Figure \ref{fig:node}. In the following sections, a detailed description of each level is provided.
The bottom layer is the hardware platform, which may be any kind of programmable microcontroller, that integrates sensors to take care of data acquisition, one or more processing elements, to manage housekeeping and pre-processing, and an adequate set of communication peripherals, implementing transmission to the gateway. \par 
The hardware platform is managed at runtime by a firmware/middleware level, potentially including some operating system (OS) support, to enable the management and scheduling of software threads. Moreover, this level must expose a set of low-level primitives to control hardware architecture details (e.g. access to peripherals, frequency, power operating mode, performance counting, etc.), and a set of monitoring Application Programming Interfaces (APIs) to continuously control the status of the hardware platform (e.g. energy and power status, remaining battery lifetime) and to characterize the performance of the different application tasks on it. \par
At the top of the node structure, there is the software application level, which executes tasks designed according to an adequate application model based on process networks, to be easily characterized and dynamically changed at runtime.\par
To implement adaptivity, we add to the application an additional software agent, that we call ADAM (ADAptive runtime Manager), which is in charge of monitoring all the events that may trigger operating mode changes (workload changes, battery status, commands from the cloud) and reconfigures the process network accordingly, to minimize power/energy consumption. Reconfiguration actions may involve changes in the process network topology (activation/deactivation of tasks and restructuring of the inter-task connectivity) and playing with the power-relevant knobs exposed by the architecture (e.g. clock frequency, power supply, supply voltage).
As mentioned, to assess the feasibility of our approach based on dynamic reconfiguration, we have used a single-core microcontroller, namely an off-the-shelf platform designed by STMicroelectronics named SensorTile. In the following, we will describe the main features of such platforms, exploited in this work.

We have chosen SensorTile to represent a class of platforms available on the market equipped with a single-core low-power IoT nodes, usually integrating a wide scope of sensors and peripherals to increase usability. These solutions often integrate mid- to low-end processing elements, capable of executing simpler near-sensor processing tasks on a low energy budget, using optimized libraries to recover performance and lightweight operating systems to enable the coexistence of multiple software processes.  
\subsection{Hardware platform layer}
The SensorTile measures $13.5 \times 13.5 mm$. It's equipped with an ARM Cortex-M4 32-bit low-power microcontroller. The small size and low power consumption allow the device to be powered also by the battery and obtaining good results in terms of autonomy without having to give up portability.
Several architectural knobs can be used to adapt the platform to different conditions. SensorTile can work in two main modalities: \emph{run mode} and \emph{sleep mode}, in which different subsets of the hardware components are active. Moreover, in each mode, the chip can be set to a different system frequency (from 0.1 MHz to 80 MHz). Depending on the chosen system frequency and operating state, the device uses different voltage regulators to power the chip.

In Table \ref{tab:STmicro}, we list some configurations selectable using the mode-management APIs offered by the platform vendor.

\begin{table}[h]
    \centering
	\small
	\begin{tabular}{L{38mm} C{20mm}}
		\hline
		\\[-2.5mm]
		RUN (Range 1) at $80\,M\!H\!z$ & $120\,\mu A/\,M\!H\!z$ \\
		\\[-2.5mm]
		RUN (Range 2) at $26\,M\!H\!z$ & $100\,\mu A/\,M\!H\!z$ \\
		\\[-2.5mm]
		LPRUN at $2\,M\!H\!z$ & $112\,\mu A/\,M\!H\!z$ \\
		\\[-2.5mm]
		SLEEP at $26\,M\!H\!z$ & $35\,\mu A/\,M\!H\!z$ \\
		\\[-2.5mm]
		LPSLEEP at $2\,M\!H\!z$ & $48\,\mu A/\,M\!H\!z$ \\
		\\[-2.5mm]
		\hline
		\\[-1mm]
	\end{tabular}
    \caption{SensorTile current consumption in different operating states.}
	\label{tab:STmicro}
\end{table}

For our experiments, we have chosen to use two approaches to dynamically reduce power consumption: 
\begin{itemize}
    \item to change system frequency (and consequently voltage regulator settings) over time according to the workload.
    \item to use the sleep mode of the microcontroller whenever possible. The operating system automatically sets a sleep state when there are no computational tasks queued to be performed and a timer-based awakening can be used to restart the run mode when needed.
\end{itemize}

\subsection{Middleware/OS layer}
In addition to the API offered by the manufacturer, we used other middleware components to manage multiple computation tasks at runtime and to execute CNN-based near-sensor processing with an adequate performance level. 
\subsubsection{FreeRTOS}
SensorTile runs FreeRTOS as ROTS (Real-time Operating System). This firmware component is aimed at developers who intend to have a real-time operating system without too much impact on the memory footprint of the application. The size of the operating system is between $4\,kB$ and $9\,kB$. Some features offered by the operating system are real-time scheduling functionality, communication between processes, synchronization, time measurements. One of the most important aspects that led us to choose FreeRTOS is that of having the possibility to enable thread-level abstraction to represent processing tasks to be executed on the platform and to timely manage their scheduling at runtime.
 
FreeRTOS creates a system task called \textit{idle} task, which is set with the lowest possible execution priority. When this task is executed, the system tick counter is deactivated and the microcontroller is put in a sleep state. Due to the priority setting, the idle task is only executed if there are no other tasks waiting to be called by the scheduler.

FreeRTOS does not natively support the frequency variation of the system. Once the frequency has changed, timing functions would be completely de-synchronized. We had to modify part of the OS support, to enable system frequency changes without impact on the rest of the OS functionality.


\subsubsection{CMSIS}
In order to be capable of executing in-place processing of the sensed data, we have exploited the Cortex Microcontroller Software Interface Standard (CMSIS), an optimized library specifically targeting Cortex-M processor cores \cite{Lai2018}.
It includes several modules having many libraries capable of optimizing mathematical functions based on the type of architecture used. Of particular interest is the CMSIS-NN module, inside there are various optimized functions that allow cognitive computational implementations.
While CMSIS provides quite extensive support for neural network execution, we had to add some changes to support the use-case that are described in the following, namely to enable mono-dimensional convolutions on one-dimension sensor data streams.

\subsection{Application model}
\label{sec:app_model}
In this section, we describe the application model that we have used to create and analyze the application, the source code is available at our public repository\footnote{\url{https://github.com/matteoscrugli/adam-iot-node-on-stm32l4}}. We selected an application structure based on process networks. Tasks are represented as independent processes, communicating with each other via FIFO structures, using blocking \emph{read} and \emph{write} communication primitives to avoid data loss in case of busy pipeline stages. Processes may be potentially executed in parallel, in case of available processing resources, potentially improving performance using a software pipeline.  

\begin{figure}[h]
	\tiny
	\centering
	\begin{tikzpicture}[scale=0.33,->,>=stealth',initial text={},shorten >=1pt,auto,node distance=1.7cm,semithick]
	\node[state, minimum size=1.2cm] (x0)					    {\textit{Get data}};
	\node[state, minimum size=1.2cm] (x1) [right of=x0]   {\textit{Process}};
	\node[state, minimum size=1.2cm] (x2) [right of=x1]   {\textit{Threshold}};
	\node[state, minimum size=1.2cm] (x3) [right of=x2] 	{\textit{Send}};
	
	\path
	(x0)	edge                    node {\textit{$ $}}			(x1)
	(x1)	edge            		node {\textit{$ $}}	 		(x2)
	(x2)	edge            		node {\textit{$ $}}	 		(x3);
	\end{tikzpicture}
	\caption{Simple task chain.}
	\label{fig:thread_baseconf}
\end{figure}
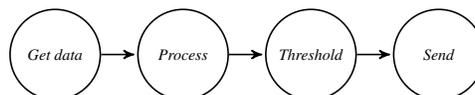

In particular, for each sensed variable to be monitored, we build a chain of tasks that operate on the sensed data (Figure \ref{fig:thread_baseconf}).

A chain of processes is generated for each sensor node, so that, if required by changes in the operating mode, it's possible to dynamically turn on and off the useful and non-useful components.

For each sensor, we envision four types of general tasks:
\begin{itemize}
\item \textit{Get data task:} takes care of taking data from the sensing hardware integrated into the node.
\item \textit{Process task:} it's possible to have multiple tasks of this type, representing multiple stages of in-place data analysis algorithm. Having more than one task of this type allows a prospective user to select, for example, a certain depth of analysis, which determines an impact on the required communication bandwidth, detail of the extracted information, and power/energy consumption.
\item \textit{Threshold task:} this task allows to filter data depending on the results of the in-place analysis. For example, a threshold task may be used to send data to the cloud only when specific events or alert conditions are detected. Its purpose is to limit data transfers from the node. 
\item \textit{Send task:} is the task in charge of outwards communication to the gateway.
\end{itemize}

Considering the selected process network model, activation/deactivation of tasks or entire chains corresponding to sensors can be implemented by: 
\begin{itemize}
    \item enabling/stopping the periodic execution of the involved task;
    \item reconfiguring the FIFOs to reshape the process chain accordingly.
\end{itemize}
In this way, it's possible to select multiple application configurations, corresponding to operating modes characterized by different levels of in-place computing effort, bandwidth requirements, monitoring precision.\par 

\subsection{Adaptivity support: the ADAptive runtime Manager}
\label{sec:adam}
Within the process network, a task was exclusively dedicated to the management of dynamic hardware and software reconfiguration of the platform. We have implemented such reconfiguration in a software agent called ADAptive runtime Manager (ADAM). ADAM can be activated periodically by means of an internal timer. It evaluates the status of the system, monitoring:
\begin{itemize}
    \item reconfiguration commands from the gateway; 
    \item changes in the workload, e.g. rate of events to be processed. For example, a task may have to be executed periodically, with a rate that depends on the frequency of certain events in the sensor data. This poses real-time constraints that may be varying over time in a data-dependent manner. 
    \item other relevant variables (e.g. battery status).
\end{itemize}

Depending on such input, ADAM can react to change the platform settings, performing different operations:  
\begin{itemize}
\item Enable or disable the individual tasks of the sensor task chain or the entire chain;
\item Choose whether to set the microcontroller in a sleep mode or not;
\item Set the operating frequency of the microcontroller to increase/reduce performance level;
\item Reroute the data-flow managed by the FIFOs according to the active tasks.
\end{itemize}

\begin{figure}[h]
	\tiny
	\centering
	\begin{tikzpicture}[scale=0.33,->,>=stealth',initial text={},shorten >=1pt,auto,node distance=1.7cm,semithick]
	\node[state, minimum size=1.2cm] (x0)					    {\textit{Get data}};
	\node[state, minimum size=1.2cm] (x1) [right of=x0]   {\textit{Process}};
	\node[state, minimum size=1.2cm] (x2) [right of=x1]   {\textit{Threshold}};
	\node[state, minimum size=1.2cm] (x3) [right of=x2] 	{\textit{Send}};
	
	\path
	(x0)	edge                    node {\textit{$ $}}			(x1)
	    	edge [bend left=37, ggrigio, dashed]     node {\textit{$ $}}		    (x3)
	(x1)	edge            		node {\textit{$ $}}	 		(x2)
	(x2)	edge            		node {\textit{$ $}}	 		(x3);
	\end{tikzpicture}
	
	\begin{tikzpicture}[scale=0.33,->,>=stealth',initial text={},shorten >=1pt,auto,node distance=1.7cm,semithick]
	\node[state, minimum size=1.2cm] (x0)					    {\textit{Get data}};
	\node[state, minimum size=1.2cm, draw=ggrigio, text=ggrigio] (x1) [right of=x0]   {\textit{Process}};
	\node[state, minimum size=1.2cm, draw=ggrigio, text=ggrigio] (x2) [right of=x1]   {\textit{Threshold}};
	\node[state, minimum size=1.2cm] (x3) [right of=x2] 	{\textit{Send}};
	
	\path
	(x0)	edge [ggrigio, dashed]		      node {\textit{$ $}}			(x1)
	    	edge [bend left=37]           	  node {\textit{$ $}}		    (x3)
	(x1)	edge [ggrigio, dashed]		      node {\textit{$ $}}	 		(x2)
	(x2)	edge [ggrigio, dashed]		      node {\textit{$ $}}	 		(x3);
	\end{tikzpicture}
	\caption{Two possible configurations of a generic system.}
	\label{fig:thread_conf}
\end{figure}
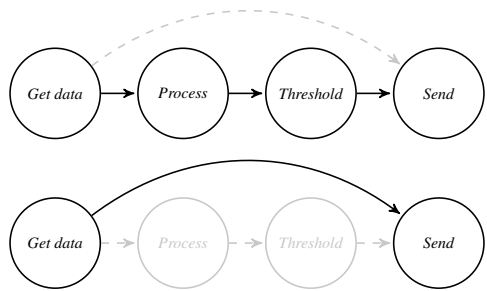

Figure \ref{fig:thread_conf} shows an example of the reconfiguration of the system that may be applied by ADAM, deactivating a \emph{process} task, to switch from an operating mode that sends pre-processed information to the cloud to another sending raw data.

\section{Designing the application: operating modes and processing tasks}
\label{sec:usecase}
To implement ECG monitoring, we have declined the previously described application model to deploy an adequate waveform analysis application on SensorTile. We built a  prototype using an AD8232 sensor module from Analog Devices, connected to the ADC converter integrated into the reference platform. In this section, we describe the supported operating modes, that can be selected at runtime, and the processing tasks coexisting in the different operating modes. 
\subsection{Operating modes}
\begin{figure}[h]
	\tiny
	\centering
	\begin{tikzpicture}[->,>=stealth',initial text={},shorten >=1pt,auto,node distance=1.7cm,semithick]
	\node
	    [state, minimum size=1.2cm] (x0)					                          {\textit{Get data}};
	\node
	    [state, minimum size=1.2cm, draw=ggrigio, text=ggrigio] (x1)
	    [right of=x0]
	    {\textit{Peak}};
	\node
	    [state, minimum size=1.2cm, draw=ggrigio, text=ggrigio] (x2)
	    [right of=x1]
	    {\textit{CNN}};
	\node
	    [state, minimum size=1.2cm, draw=ggrigio, text=ggrigio] (x3)
	    [right of=x2]
	    {\textit{Threshold}};
	\node
	    [state, minimum size=1.2cm] (x4)
	    [right of=x3]
	    {\textit{Send}};
	\node
	    [text width=5cm] at (1.75,1.48) 
        {\footnotesize Operating mode 1: Raw data.};
	\node
	    [text width=4cm] at (1.75,-1.5) 
        {};
	
	\path
	(x0)	edge [ggrigio, dashed]	node {\textit{$ $}}	(x1)
	    	edge [bend left=30]                     node {\textit{$ $}} (x4)
	(x1)	edge [ggrigio, dashed]		            node {\textit{$ $}}	(x2)
	        edge [bend right=37, ggrigio, dashed]   node {\textit{$ $}} (x3)
	(x2)	edge [ggrigio, dashed]		            node {\textit{$ $}}	(x3)
	(x3)	edge [ggrigio, dashed]                  node {\textit{$ $}}	(x4);
	\end{tikzpicture}
	
	\begin{tikzpicture}[->,>=stealth',initial text={},shorten >=1pt,auto,node distance=1.7cm,semithick]
	\node
	    [state, minimum size=1.2cm] (x0)					                          {\textit{Get data}};
	\node
	    [state, minimum size=1.2cm] (x1)
	    [right of=x0]
	    {\textit{Peak}};
	\node
	    [state, minimum size=1.2cm, draw=ggrigio, text=ggrigio] (x2)
	    [right of=x1]
	    {\textit{CNN}};
	\node
	    [state, minimum size=1.2cm] (x3)
	    [right of=x2]
	    {\textit{Threshold}};
	\node
	    [state, minimum size=1.2cm] (x4)
	    [right of=x3]
	    {\textit{Send}};
	\node
	    [text width=5cm] at (1.75,1.48) 
        {\footnotesize Operating mode 2: Peak detection.};
	\node
	    [text width=4cm] at (1.75,-1.5) 
        {};
	
	\path
	(x0)	edge                                    node {\textit{$ $}}	(x1)
	    	edge [bend left=30, ggrigio, dashed]    node {\textit{$ $}} (x4)
	(x1)	edge  [ggrigio, dashed]   		        node {\textit{$ $}}	(x2)
	        edge [bend right=37]                    node {\textit{$ $}} (x3)
	(x2)	edge  [ggrigio, dashed]   		        node {\textit{$ $}}	(x3)
	(x3)	edge                                    node {\textit{$ $}}	(x4);
	\end{tikzpicture}
	
	\begin{tikzpicture}[->,>=stealth',initial text={},shorten >=1pt,auto,node distance=1.7cm,semithick]
	\node
	    [state, minimum size=1.2cm] (x0)					{\textit{Get data}};
	\node
	    [state, minimum size=1.2cm] (x1)
	    [right of=x0]
	    {\textit{Peak}};
	\node
	    [state, minimum size=1.2cm] (x2)
	    [right of=x1]
	    {\textit{CNN}};
	\node
	    [state, minimum size=1.2cm] (x3)
	    [right of=x2]
	    {\textit{Threshold}};
	\node
	    [state, minimum size=1.2cm] (x4)
	    [right of=x3]
	    {\textit{Send}};
	\node
	    [text width=5cm] at (1.75,1.48) 
        {\footnotesize Operating mode 3: CNN processing.};

	\path
	(x0)	edge                                    node {\textit{$ $}}	(x1)
	    	edge [bend left=30, ggrigio, dashed]    node {\textit{$ $}} (x4)
	(x1)	edge                    		        node {\textit{$ $}}	(x2)
	        edge [bend right=37, ggrigio, dashed]   node {\textit{$ $}} (x3)
	(x2)	edge                                    node {\textit{$ $}}	(x3)
	(x3)	edge                                    node {\textit{$ $}}	(x4);
	\end{tikzpicture}
	
	\caption{EEG application model.}
	\label{fig:opmodeconf}
\end{figure}
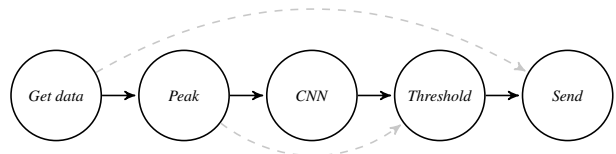

We have enabled three different operating modes to be selectable by the user, by sending adequate commands from the cloud. Operating modes are shown in Figure \ref{fig:opmodeconf}.

\subsubsection{Operating mode 1: Raw data}
The first operating mode envisions sending the entire data stream acquired by the sensor node to the gateway. There is therefore no near-sensor data analysis enabled, and it poses fairly high requirements in terms of bandwidth. 
In this operating mode are:
\begin{itemize}
\item Multiple samples are been grouped and inserted into a packet of 20 Bytes (8 ECG data 16 bit, 1 timestamp 32 bit).
\item The sample rate of the ADC is set to $330\,H\!z$, considering sending multiple samples at a time, one Bluetooth packet is sent every $24\,ms$.
\end{itemize}

\subsubsection{Operating mode 2: Peak detection}
This operating mode does not provide visual access to the whole ECG waveform. A healthcare practitioner, when selecting this mode when accessing the data, can select to monitor only heartbeat rate, requiring a lower level of detail in the information sent to the cloud. He could also set thresholds and receive notification
only when thresholds are exceeded. 
In this operating mode, four tasks are active:
\begin{itemize}
\item Get data task
\item Process data (peak detection)
\item Threshold task (alert heartbeat rate evaluation)
\item Send task
\end{itemize}
This operating mode processes samples to search for signal peaks and consequently computes the heartbeat rate. The first task (Figure \ref{fig:opmodeconf}) collects data from the sensor (as in raw data operating mode), the second analyzes the signal analysis and calculates the heart rate, and the fourth allows data transmission.
The threshold task is used to determine if data must be sent to the cloud. For example, no data is sent if the heartbeat rate is controlled between two high and low alert values.
The peak detection algorithom it's not very critical in terms of time and power consumption, it will be better discussed in Section \ref{sec:peakdetection}. 
The size of the package sent is 5 Bytes packet (1 heartbeat rate value, represented on 8 bit, 1 time-stamp 32 bit). The transmission rate is given dependent, in the worst case a package is sent for each peak detected.

\subsubsection{Operating mode 3: CNN processing}
In the latter operating mode, a further level of analysis is introduced. An additional task implements a convolutional neural network, classifying the ECG waveform to recognize physically relevant conditions. Using such classification technique, the practitioner can monitor the morphology of the signal without the need of sending the entire data stream to the cloud, saving transmission-related power/energy consumption.
The neural network implemented recognizes anomalous occurrences in the ECG tracing, in this case, communications with the gateway occur only in case of anomaly detection.
The enabled tasks are:
\begin{itemize}
\item Get data task
\item Process data 1 task (peak detection)
\item Process data 2 task (CNN)
\item Threshold task (anomalous shapes in the ECG waveform)
\item Send task
\end{itemize}
The required communication bandwidth is more similar to \textit{peak detection o.m.} than \textit{raw data o.m.}, however, with respect to \textit{peak detection o.m.}, computing effort is higher. The node executes the 1D convolution neural network similar to the one described in \cite{cnn}.
We have designed the system be capable of classifying ECG peaks according to alternative sets of categories, each composed by 5 classes,  named \textit{NLRAV} and \textit{NSVFQ} (see Figure \ref{fig:cnn}).
The design process used to select, train and deploy the specific neural network topology is explained in Section \ref{sec:training}.


The size of the data transferred to the cloud is 6 Bytes (1 heartbeat data 8 bit, 1 label data 8 bit, 1 timestamp 32 bit).

\subsection{The peak detection algorithm}
\label{sec:peakdetection}
\begin{figure}
  \centering
    \begin{center}
        \begin{tikzpicture}
            \scriptsize
            \sbEntree{in}
            
            \sbBloc [4]{a}{\textit{DC filter}} {in}
            \sbBlocL{b}{\textit{LP filter}} {a}
            \sbBlocL{c}{\textit{Derivative}}{b}
            \sbBlocL{d}{\textit{Squared}}   {c}
        
            \sbSortie[4]{out}{d}
            
            \sbRelier[\textit
            {
                \begin{tabular}{c}
                    Raw \\
                    signal \\
                \end{tabular}
            }
            ]{in}{a}
            \sbRelier[\textit
            {
                \begin{tabular}{c}
                    Filtered \\
                    signal \\
                \end{tabular}
            }
            ]{d}{out}
        
        \end{tikzpicture}
    \end{center}
  \caption{Filtering block diagram.}
  \label{fig:bd}
\end{figure}
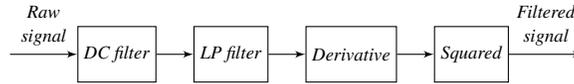

The processing of the ECG signal is activated in \textit{peak detection} and \textit{CNN o.m.}, in both operating modes it's necessary to identify the R peaks in the signal, therefore a simplified version of the Pan Tompkins algorithm was used in order to obtain the position of the R peaks during data acquisition from the sensor. The reference study to implement the R peak recognition algorithm is \cite{pt}, the Figure \ref{fig:bd} shows the block diagram representing the signal processing.

\begin{figure}
  \centering
  \includegraphics[width=0.65\linewidth]{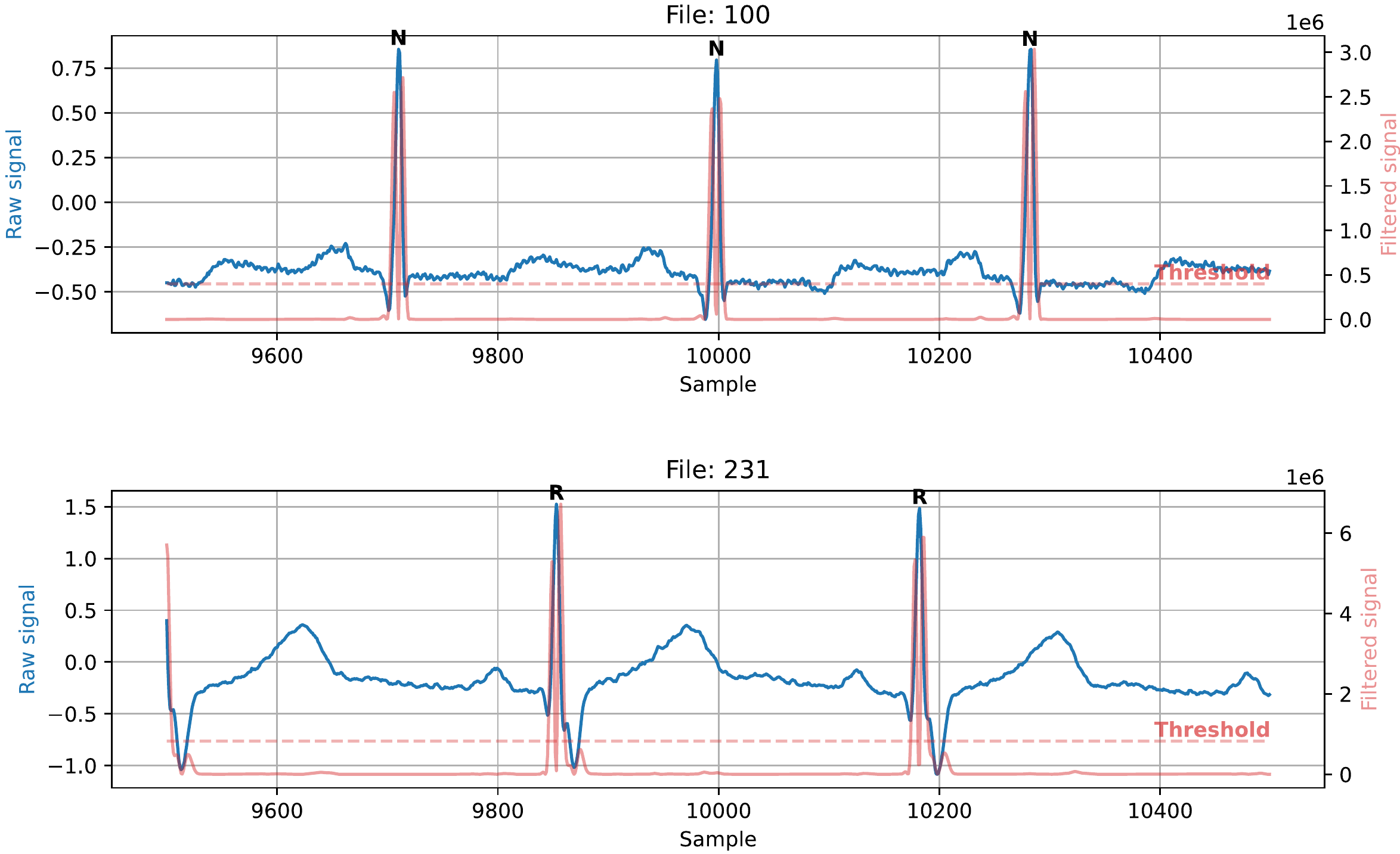}
  \caption{Raw signal and filtered one from two different recording.}
  \label{fig:signal}
\end{figure}

Figure \ref{fig:signal} shows the raw signal in blue color and the filtered one in red from two different recordings. A peak is detected when a filtered signal exceeds a predefined threshold, then returns to a local minimum point and the delay introduced by the filter is taken into account, the threshold value may be set differently for each recording. 

A detected peak is considered a true positive when it is associated with a dataset peak in a neighborhood of 50 samples within the track under analysis. Equation~\ref{eq:accpeak50_0} and \ref{eq:accpeak50_1} shows the sensitivity (true positive rate) and precision (positive predictive value) data of the peak detection algorithm on the MIT-BIH arrhythmia database:
\begin{gather}
    \label{eq:accpeak50_0}
    \textit{TPR} = \frac{\textit{TP}}{\textit{TP} + \textit{FN}} = 0.99674\,,\\
    \label{eq:accpeak50_1}
    \textit{PPV} = \frac{\textit{TP}}{\textit{TP} + \textit{FP}} = 0.99421\,.
\end{gather}


\begin{figure}
    \centering
        \includegraphics[width=0.4\columnwidth]{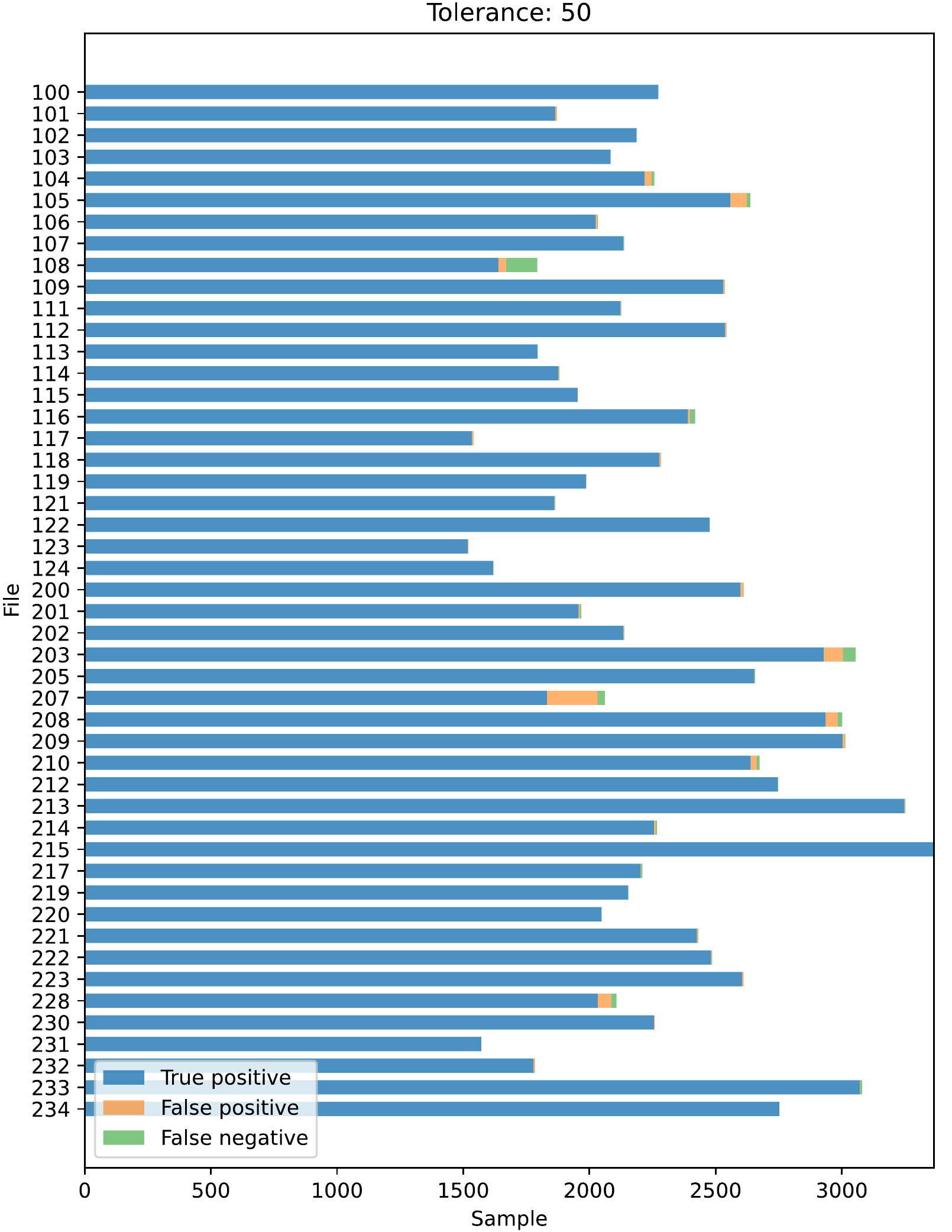}
    \caption{True positive, false positive and false negative identified by the peak detection algorithm.}
    \label{paolo}
\end{figure}


Figure \ref{paolo} shows true positive, false positive, false negative of the peak detection algorithm with a tolerance of 50 samples.


\subsection{Designing the CNN: training and optimization}
\label{sec:training}
We have exploited a training procedure using and comprising a static quantization\footnote{\label{note:quant}\url{https://pytorch.org/tutorials/advanced/static_quantization_tutorial.html}} step, the source code is available at our public repository\footnote{\url{https://github.com/matteoscrugli/ecg-classification-quantized-cnn}}. This process enables the conversion of weights and activations from floating-point to integers and allows to implementation of the CNN using the CMSIS-NN optimized function library, which expects inputs represented with 8-bit precision.
In static quantization\textsuperscript{\ref{note:quant}}, which takes place right after quantization, \textit{float} values are converted to \textit{qint8} format. We set the procedure to force bias values to be null, while, to quantize weights,  \textit{MinMax} observers\footnote{\url{https://pytorch.org/docs/stable/_modules/torch/quantization/observer.html}} are inserted inside the network to detect the output values dynamics in each layer. On the basis of the reported distribution, \textit{scale} and \textit{zero-point} values are selected and used to convert effectively and prevent data saturation. 

The functions implementing convolution and fully connected layers in the CMSIS-NN library provide for output shifting operations to apply the \textit{scale} factor on the outputs, allowing for scaling values ranging from -128 to 127. The quantization procedure in PyTorch, on the other hand, requires a \textit{scale} value that is not necessarily a power of 2. For this reason, we slightly modified the CMSIS functions to support arbitrary \textit{scale} values. Such modification has led to a limited increase of the inference execution time. As an example of such performance degradation, we report here the execution time increase for two examples CNN topologies, named \textit{20\_20\_100} and \textit{4\_4\_100} networks (network name indicates the main topology parameters as \textit{conv1OutputFeatures\_ conv2OutputFeatures\_ fc1Outputs}), corresponding to respectively 2,87\% and 10.52\%.

\subsubsection{Model exploration}
\begin{figure}
    \centering
    \begin{minipage}{0.6\textwidth}
        \centering
  \includegraphics[width=1\textwidth]{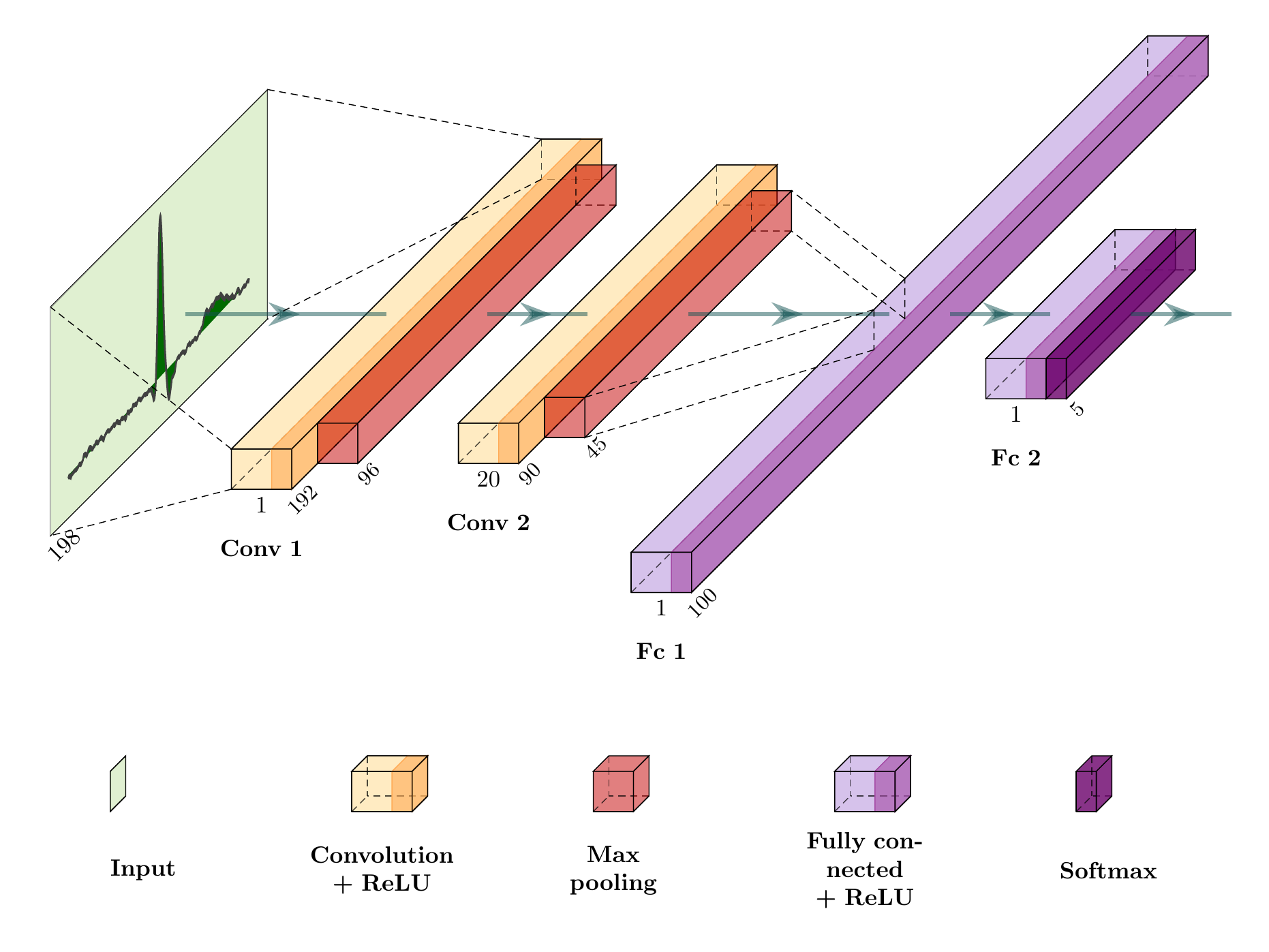}
    \end{minipage}
    \\[2.5mm]
    \begin{minipage}{0.6\textwidth}
        \centering
	\scriptsize
	\begin{tabular}{L{36mm} L{38mm}}
		\hline
		\\[-2.5mm]
		\textbf{(N)} Normal beats & \textbf{(N)} Normal beats\\
		\\[-2.5mm]
		\textbf{(L)} Left bundle branch block & \textbf{(S)} Superventricular ectopic beats \\
		\\[-2.5mm]
		\textbf{(R)} Right bundle branch block & \textbf{(V)} Ventricular premature contraction \\
		\\[-2.5mm]
		\textbf{(A)} Atrial premature contraction & \textbf{(F)} Fusion beats \\
		\\[-2.5mm]
		\textbf{(V)} Ventricular premature contraction  & \textbf{(Q)} Unclassificable beat\\
		\\[-2.5mm]
		\hline
		\\[-2mm]
	\end{tabular}
    \end{minipage}
  \caption{CNN structure and two possible classes of labels.}
  \label{fig:cnn}
\end{figure}

In order to select an optimized CNN topology implementing the classification task required for the system, we have carried out a design space exploration process, comparing tens of neural network topologies in terms of accuracy reached after training and in terms of computing workload associated with executing the inference task on SensorTile. We have explored multiple topologies composed by two convolution layers, two down-sampling layers, and two fully connected layers, as represented in Figure \ref{fig:cnn}, the size of the input sample frame is equal to 198.
 \par
\begin{figure}
    \centering
    \includegraphics[width=0.5\textwidth]{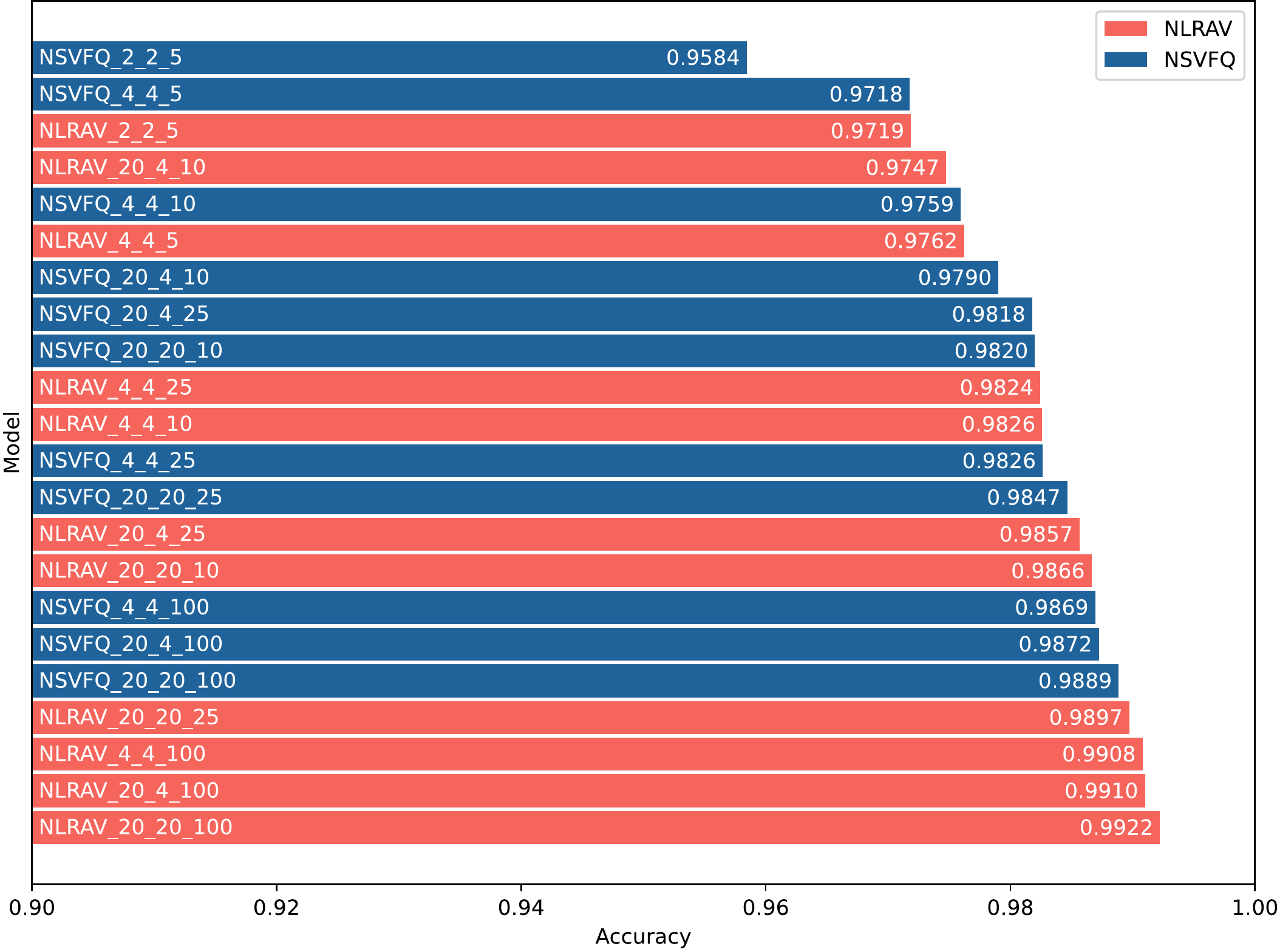}
    \caption{Exploration of the chosen neural network model, the name comes from \textit{labels\_ conv1OutputFeatures\_ conv2OutputFeatures\_ fc1Outputs}.}
    \label{fig:train_exp}
\end{figure}
\begin{figure}
    \centering
    \begin{subfigure}[b]{0.65\textwidth}
       \includegraphics[width=1\linewidth]{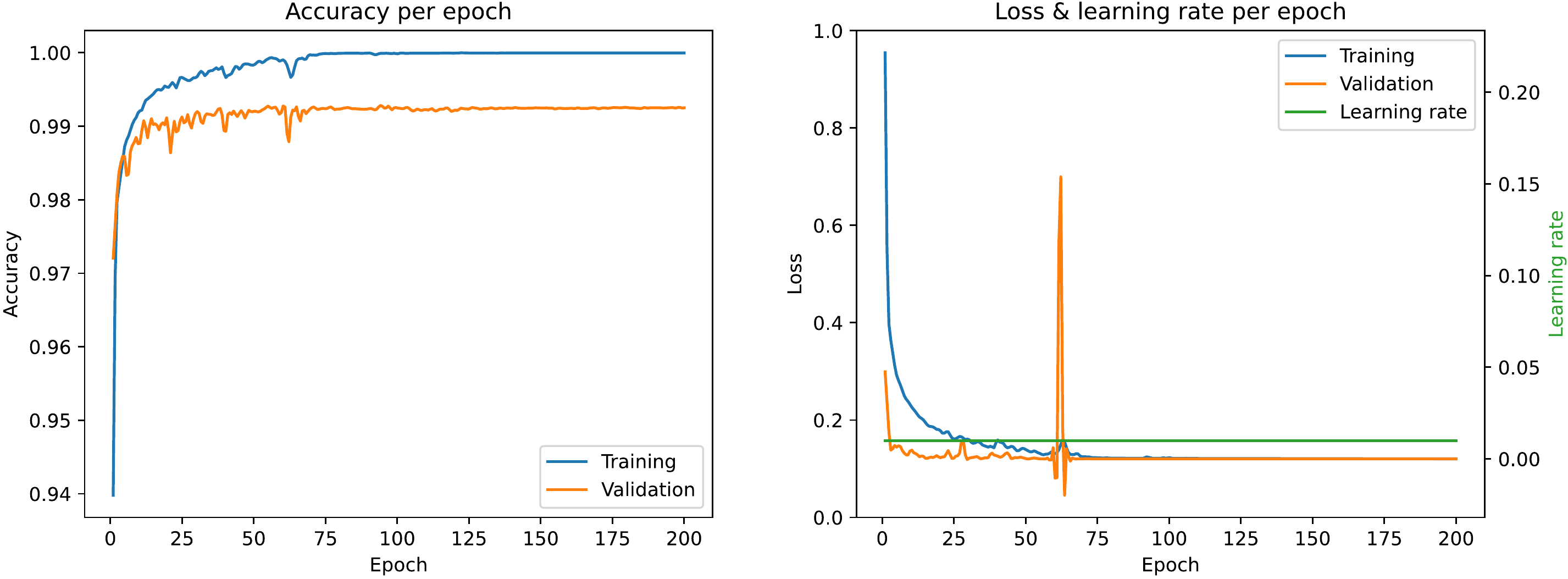}
       \caption{NLRAV classes training results.}
    \end{subfigure}
    \\[3mm]
    \begin{subfigure}[b]{0.65\textwidth}
       \includegraphics[width=1\linewidth]{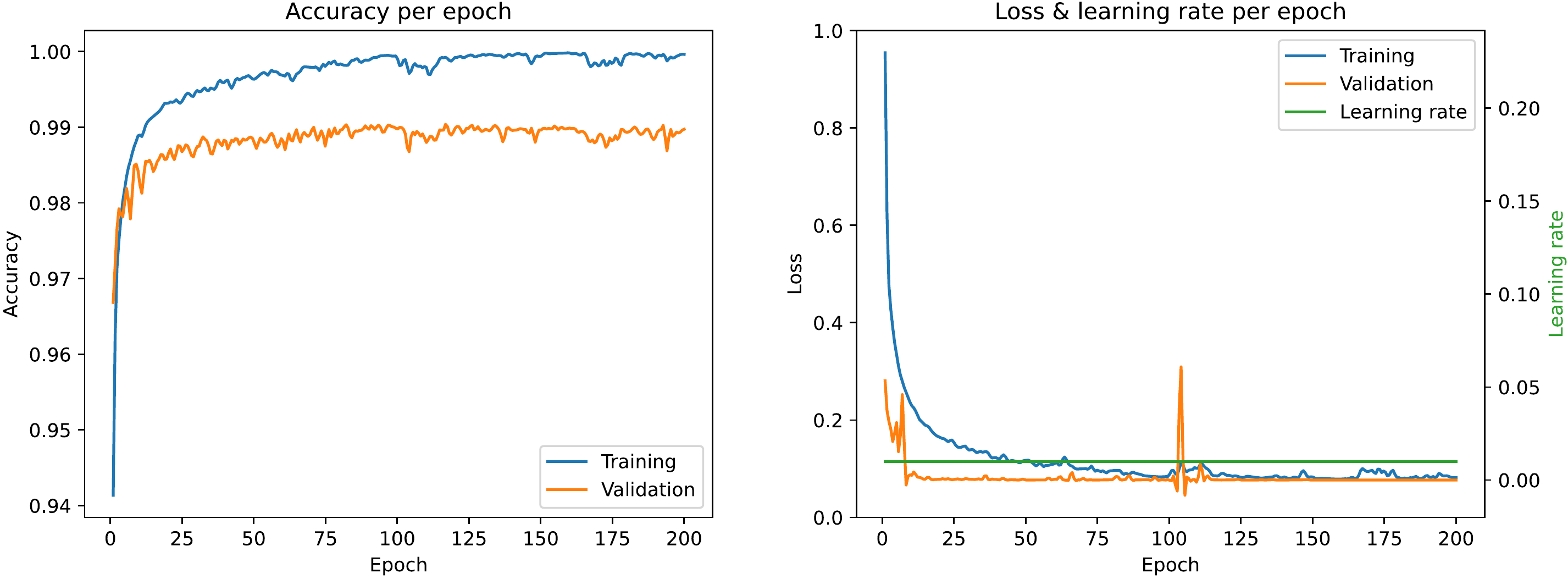}
       \caption{NSVFQ classes training results.}
    \end{subfigure}
    \caption{Results obtained from the training of the model having: 20 output features for \textit{Conv1}, 20 output features for \textit{Conv2} and 100 output for \textit{Fc1}.}
    \label{fig:train}
\end{figure}

Explored topologies feature different numbers of output channels from each layer. The results are reported in Figure \ref{fig:train_exp}, showing the most interesting results for both the \textit{NLRAV} and \textit{NSVFQ} classes. Models \textit{NLRAV\_20\_20\_100} and \textit{NSVFQ\_20\_20\_100} achieve the highest accuracy value as shown in Equation \ref{eq:mostacc_0} and \ref{eq:mostacc_1}. The training set is compoused by 70\% of the elements of the entire dataset and they are chosen randomly. Figure \ref{fig:train} shows the trend of the accuracy value during the training stage.
\begin{align}
    \label{eq:mostacc_0}
    \textit{ACC\textsubscript{NLRAV\textunderscore20\textunderscore20\textunderscore100}} &= \frac{\textit{TP}}{\textit{TP} + \textit{FP} + \textit{FN}} = 0.9922\,,\\
    \label{eq:mostacc_1}
    \textit{ACC\textsubscript{NSVFQ\textunderscore20\textunderscore20\textunderscore100}} &= \frac{\textit{TP}}{\textit{TP} + \textit{FP} + \textit{FN}} = 0.9889\,.
\end{align}

\begin{figure}
    \centering
    \includegraphics[width=0.55\textwidth]{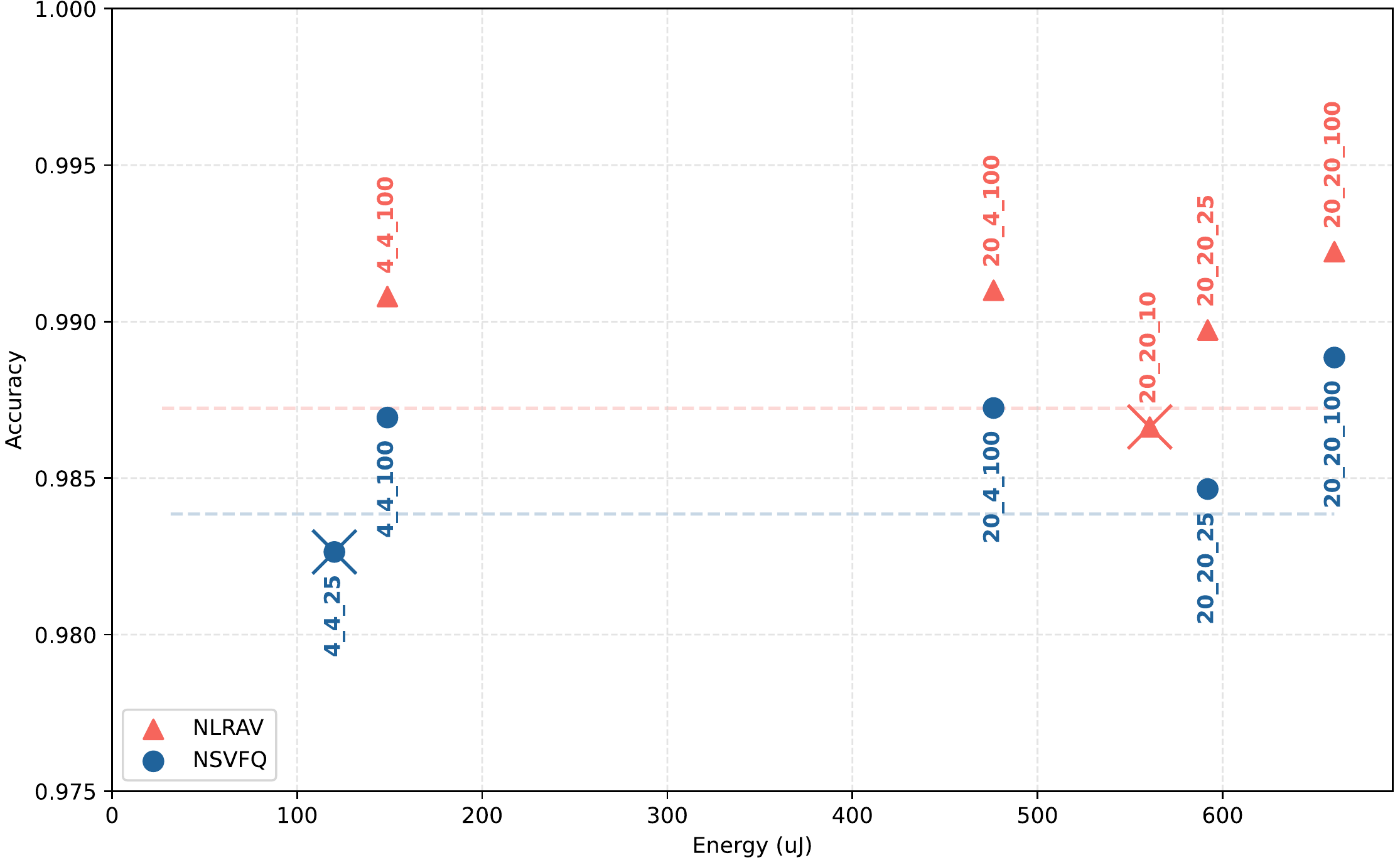}
    \caption{For the most accurate models, the energy consumption for a single CNN task call is shown. The dotted lines represents the maximum allowable drop in accuracy (0.5\% with respect to the most accurate model) for NLRAV (red line) and NSVFQ (blue line) classes. The models marked with an ``\texttimes''~do not respect the constraints imposed on the minimum necessary accuracy value.}
    \label{fig:model_sel}
\end{figure}

Figure \ref{fig:model_sel} shows a Pareto plot representing accuracy and energy consumption for the most accurate topologies identified by the exploration.

For both classes NLRAV and NSVFQ, only one neural network model must be selected which allows to reduce power consumption as much as possible but, at the same time, does not lead to an excessive drop in accuracy. A maximum accuracy drop equal to 0.5\% with respect to the most accurate model (represented in Figure \ref{fig:model_sel} by the dotted lines) was chosen. 
The reported energy consumption is associated with a single CNN inference task execution on SensorTile. Models that are above the 0.5\% threshold are considered to be valid, and, for each set of labels, the valid model that consumes less energy is chosen to be refined in the next steps and deployed on the board.
Eventually, we have selected \textit{NLRAV\_4\_4\_100} and \textit{NSVFQ\_4\_4\_100}. The accuracy values are reported in Equation \ref{eq:smallacc_0} and \ref{eq:smallacc_1}.
\begin{align}
    \label{eq:smallacc_0} 
    \textit{ACC\textsubscript{NLRAV}} &= 0.9908\,,\\
    \label{eq:smallacc_1}
    \textit{ACC\textsubscript{NSVFQ}} &= 0.9869\,.
\end{align}

\subsubsection{Post-deployment degradation and Refinement with Augmentation}
The ECG peaks in the reference dataset are perfectly centered in the frame of samples that is received in input by the CNN during the trainig stage. As a consequence, the network is trained to recognize the chosen classes as long as the peak is centered in the signal frame. The peak detection algorithm on the SensorTile, on the other hand, operates online on the incoming signals and would not always detect the peak in the same position specified in the dataset. \par
To assess the accuracy degradation after the deployment, we report post-deployment accuracy values by:
\begin{itemize}
    \item considering false positive and false negative peaks produced by the peak detection algorithm, which need to be accounted for in Equation \ref{eq:smallacc_0} and \ref{eq:smallacc_1}.
    \item using a post-deployment validation dataset, composed by the same samples in the original one, but modified to be centered as dictated by the peak detection algorithm during online analysis.
\end{itemize}
In these conditions, accuracy degrades to 94.52\% and 94.09\% for NLRAV and NSVFQ respectively. 

\begin{figure}
    \centering
    \includegraphics[width=0.55\textwidth]{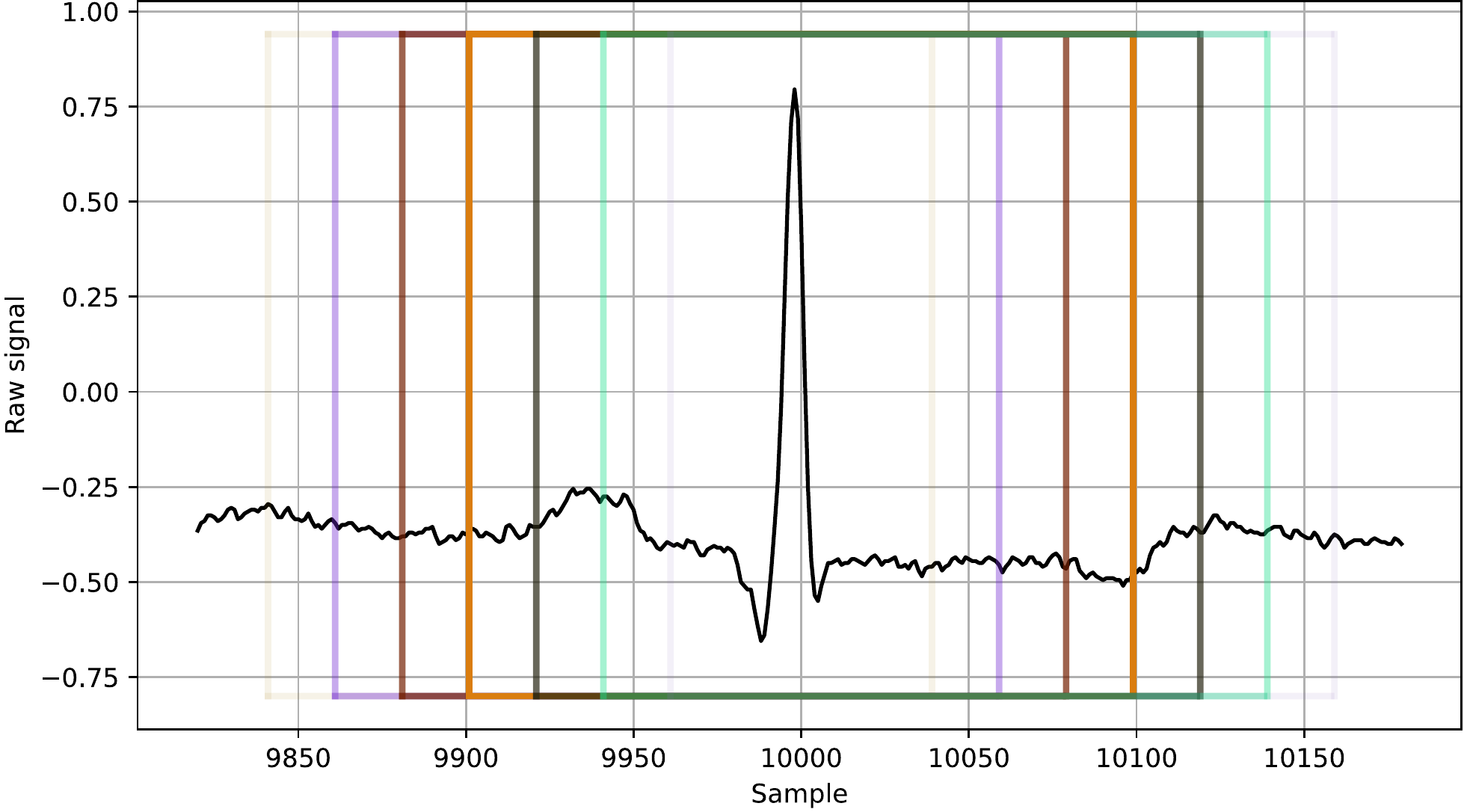}
  \caption{Qualitative example of augmentation.}
  \label{fig:aug}
\end{figure}

To overcome the deriving inaccuracy, the chosen networks have been retrained for refining their precision in case of imperfectly centered input frames. We have used a data augmentation technique, depicted in Figure \ref{fig:aug}, that adds to the training set 33 decentralized copies of each original peak in the dataset, each shifted by 3 samples with respect to the previous one.

As expectable, augmentation reduces the specialization of the CNN on the perfectly centered validation set, sightly reducing the pre-deployment accuracy to a value of 98.37\% and 97.76\% for \textit{NLRAV} and \textit{NSVFQ} respectively. On the other hand, ECG recordings with anomalous peaks, that are difficult to be perfectly centered by the peak detection algorithm, are expected to be classified much more accurately, as will be shown in Section \ref{sec:results}.

\section{Experimental results}
\label{sec:results}
In this section, we show our main experimental results. We first show a detailed accuracy evaluation to show the effectiveness of the data augmentation procedure and the class-level classification capabilities of the designed CNNs. Moreover, we present measures of the energy consumption of the entire system and we highlight the energy contributions of each task. To estimate power consumption in each operating mode, we have performed a thorough set of experiments measuring energy consumption in different setup conditions. The results were used to create a model highlighting the contribution of each task to the energy consumption of the node.

\subsection{Post-deployment CNN accuracy}
As mentioned above, when considering the ideally centered samples in the dataset, the selected CNNs are very accurate. The precision of the classification, however, decreases significantly when peaks are detected online and imperfectly centered.
As a solution to such accuracy degradation, we have enriched the training set with samples derived from the original ones by applying some artificial shifting. 
To prove the obtained improvements, we report a detailed classification analysis. In Table \ref{tab:matrix}, we report the number of false positives and false negatives cases resulting from the peak detection algorithm, and we classify the remaining cases, true positives, with the neural network selected for \textit{NLRAV} and \textit{NSVFQ} classes. Such classification is executed on the post-deployment validation set mentioned in Section \ref{sec:training}.
\\
The improvement in post-deployment accuracy after data augmentation is shown in Equations \ref{eq:postacc_0} and \ref{eq:postacc_1}.
\begin{align}
    \label{eq:postacc_0} 
    \textit{ACC\textsubscript{NLRAV\textunderscore post}} &= 0.9742\,,\\
    \label{eq:postacc_1}
    \textit{ACC\textsubscript{NSVFQ\textunderscore post}} &= 0.9698\,.
\end{align}

Data augmentation techniques allows recovering most (around 2.9\%) of the drop due to imperfect centering of the input ECG peaks. Data augmentation has obviously no effect on the drop due to misdetections, which still determine 1.7\% degradation with respect to the pre-deployment phase. 
\begin{table}
	\footnotesize
	\centering
    \begin{tabular}{cc|ccccc|c}
        \multicolumn{1}{c}{} &\multicolumn{1}{c}{} &\multicolumn{5}{c}{True} & \\[1ex]
        \multicolumn{1}{c}{} & 
        \multicolumn{1}{c|}{} & 
        \multicolumn{1}{c}{N} & 
        \multicolumn{1}{c}{L} & 
        \multicolumn{1}{c}{R} & 
        \multicolumn{1}{c}{A} & 
        \multicolumn{1}{c|}{V} & 
        \multicolumn{1}{c}{FP} \\ \cline{2-8}
        \multirow[c]{8}{*}{\rotatebox[origin=tr]{90}{Predicted}}
        
        & N  & 22174 & 8 & 5 & 59 & 51 &
        \multirow[c]{8}{*}{189} \\[1.5ex]
        & L & 41 & 2431 & 1 & 0 & 23 \\[1.5ex]
        & R & 52 & 0 & 2098 & 24 & 2 \\[1.5ex]
        & A & 135 & 2 & 12 & 626 & 7 \\[1.5ex]
        & V & 57 & 2 & 2 & 1 & 2094 \\[1.5ex]
        
        \cline{2-8}
        & FN & \multicolumn{5}{c|}{107}
    \end{tabular}
    \quad
    \begin{tabular}{cc|ccccc|c}
        \multicolumn{1}{c}{} &\multicolumn{1}{c}{} &\multicolumn{5}{c}{True} & \\[1ex]
        \multicolumn{1}{c}{} & 
        \multicolumn{1}{c|}{} & 
        \multicolumn{1}{c}{N} & 
        \multicolumn{1}{c}{S} & 
        \multicolumn{1}{c}{V} & 
        \multicolumn{1}{c}{F} & 
        \multicolumn{1}{c|}{Q} & 
        \multicolumn{1}{c}{FP} \\ \cline{2-8}
        \multirow[c]{8}{*}{\rotatebox[origin=tr]{90}{Predicted}}
    
        & N  & 26949 & 60 & 64 & 7 & 12 &
        \multirow[c]{8}{*}{189} \\[1.5ex]
        & S & 307 & 509 & 19 & 0 & 3 \\[1.5ex]
        & V & 87 & 10 & 2027 & 13 & 3 \\[1.5ex]
        & F & 57 & 2 & 14 & 169 & 1 \\[1.5ex]
        & Q & 36 & 2 & 6 & 0 & 2419 \\[1.5ex]
        
        
        \cline{2-8}
        & FN & \multicolumn{5}{c|}{107}
    \end{tabular}
    \\[3ex]
	\caption{False positives and false negatives cases resulting from the peak detection algorithm and classification with remaining true positive cases for NLRAV and NSVFQ classes using CNNs trained with augmentation techniques.}
	\label{tab:matrix}
\end{table}

\begin{figure}
    \centering
    \includegraphics[width=0.8\textwidth]{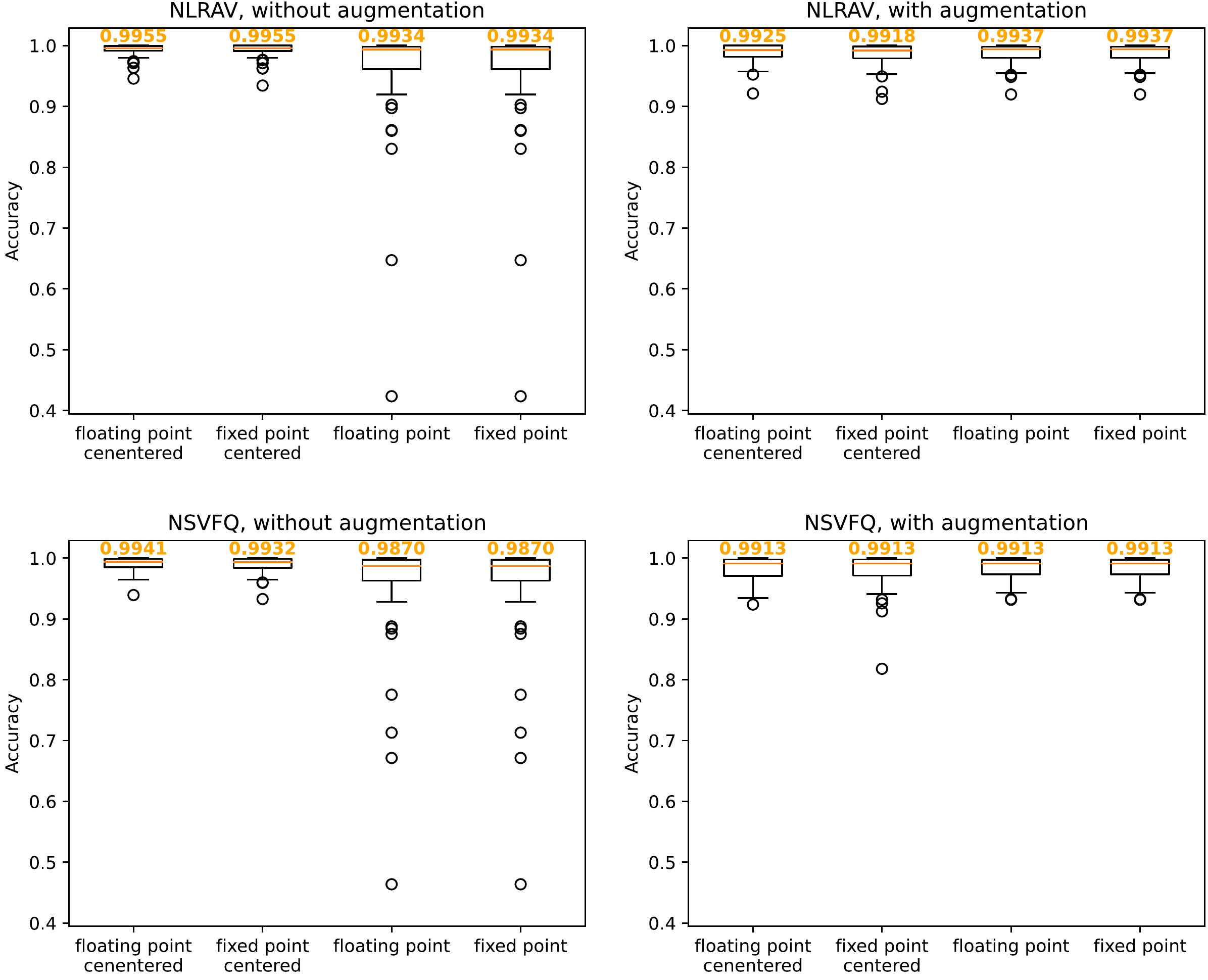}
  \caption{Taking into consideration the true positive peaks obtained with a tolerance equal to 50 samples, the statistical distribution of the accuracy values for each ECG recording, obtained from the classification on the validation set, is represented. The floating-point and fixed point models are tested, inference with centered and non-centered peaks is also tested. In orange, the median value.}
  \label{fig:tpaug_inference}
\end{figure}

Figure \ref{fig:tpaug_inference} shows a more detailed view of the effects of the quantization procedure and of the augmentation on the accuracy, focusing on the classification of the peaks detected online.
The two leftmost plots represent the accuracy levels when no augmentation is exploited. The accuracy, as can be noticed in the leftmost bar of each plot is very high, with small variability over the different tracks, and is only slightly decreased when quantization is applied to obtain a fixed-point implementation. However, when considering the positioning of the peak as identified by the online detection, as shown in the two rightmost bars of each plot, precision degrades on some of the tracks, as can be noticed by the presence of multiple outlier tracks with very bad classification accuracy. This happens independently on the data representation format since the behavior is similar for both the fixed- and floating-point implementations. 
The two graphs on the right show the impact of data augmentation. As may be noticed by the rightmost bars in these two plots, general accuracy is significantly improved: classification works correctly for all the tracks and even the outliers show an accuracy higher than 90\%.

Finally, Table \ref{tab:acc} summarizes the results in terms of neural network accuracy on MIT-BIH dataset. We compare to the works discussed in Section \ref{sec:relatedwork} that deals with inference at-the-edge. As may be noticed, our system provides accuracy that is higher or very close than the alternatives, despite being, to the best of our knowledge, the only work actually evaluating post-deployment accuracy, and considering all the contributions to errors deriving by all the steps in the online processing system.

\begin{table}
	\centering
	\footnotesize
	\begin{tabular}{C{6mm} C{10mm} C{10mm} C{10mm} C{28mm}}
		\hline
		\\[-1.6mm]
		\textit{\textbf{Work}} & \textit{\textbf{Accuracy}} & \textit{\textbf{Sensitivity}} & \textit{\textbf{Precision}}  & \textit{\textbf{Diseases}}\\
		\\[-2.5mm]
		\hline
		\hline
		\\[-0.5mm]
		
		
		
		
		
		
		\textit{\textbf{\cite{ieee6}}} &
		$ 95.98\% $ &
		$ - $ &
		$ 95.9\% $ &
		NSVF \\
		\\[0mm]
		
		\textit{\textbf{\cite{ieee5}}} &
		$ 96\% $ &
		$ - $ &
		$ - $ &
		NSVFQ \\
		\\[0mm]
		
		\textit{\textbf{\cite{rasp}}} &
		$ 96\% $ &
		$ - $ &
		$ - $ &
		[1] \\
		\\[0mm]
		
		\textit{\textbf{\cite{ieee3}}} &
		$ 97\% $ &
		$ - $ &
		$ - $ &
		NLRAV \\
		\\[0mm]
		
		\textit{\textbf{Our}} &
		$ 97.42\% $ &
		$ 98.26\% $ &
		$ 98.28\% $ &
		NLRAV \\
		\\[0mm]
		
		\textit{\textbf{Our}} &
		$ 96.98\% $ &
		$ 98.22\% $ &
		$ 98.52\% $ &
		NSVFQ \\
		\\[0mm]
		
		\hline
	\end{tabular}
	\\[2mm]
	\textsuperscript{[1]} Normal (NOR), Left Bundle Brunch Block (LBB), Right Bundle Brunch Block (RBB), Paced beat (PAB), Premature Ventricular Contraction(PVC), Atrial Premature Contraction (APC), Ventricular Flutter Wave (VFW) and Ventricular Escape Beat (VEB).
	\\[3mm]
	\caption{Results in terms of accuracy value on MIT-BIH dataset (see classes names in Figure \ref{fig:cnn}).}
	\label{tab:acc}
\end{table}

\subsection{Power consumption measures}
\begin{figure}
    \centering
    \includegraphics[width=0.485\textwidth]{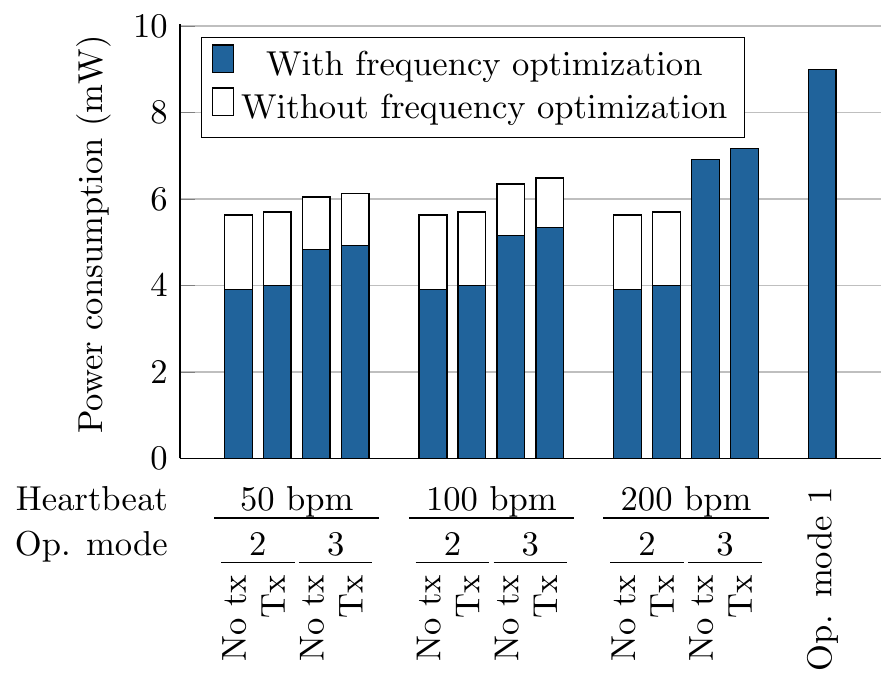}
	\caption{The graph summarizes the energy consumption for different heartbeat rates, when data sending is enabled (Tx) or not (No Tx). \textit{Raw o.m.} does not depend on heartbeat nor on the threshold settings and the threshold task is disabled, so only one value is shown.}
	\label{fig:energy}
\end{figure}

In order to measure the device's power consumption, we monitored the current absorption through an oscilloscope and a Shunt resistor. The Figure \ref{fig:energy} shows some data on power consumption derived from experimental results, the individual cases will then be taken and discussed.
\subsubsection{Case: 50 \textit{bpm}}
With low heart rates values, considerable energy savings are obtained even without adapting the system frequency to the workload. In fact, \textit{peak detection o.m.} and \textit{CNN processing o.m.} are workload-dependent, which in this case is low. For the latter reason, they consume less than the \textit{raw data o.m.}, which constantly sends data to the cloud.
There is a further energy saving given by the reduction of the system frequency according to the workload, in this case the \textit{peak detection o.m.} is set to $2\,M\!H\!z$ and the \textit{CNN processing o.m.} is set to $4\,M\!H\!z$. The \textit{raw data o.m.}, the worst case, works always at $8\,M\!H\!z$. The Figure \ref{fig:energy} also shows the power consumption values if data transmission to the cloud is present or not (Tx, No tx), as already said, the decision is up to the threshold task.
\subsubsection{Case: 100 \textit{bpm}}
\textit{Peak detection o.m.} and \textit{CNN processing o.m.} keep the same operating frequency of the previous case. Thus there is a slight increase of power consumption for such modes, only due to the more intense data-dependent workload. Obviously, no change in \textit{raw data o.m.} in terms of power consumption.
\subsubsection{Case: 200 \textit{bpm}}
Compared to the previous cases, again, there are no changes in the \textit{raw data o.m.}. A increase of the working frequency to $8\,M\!H\!z$ is required to sustain the \textit{CNN processing o.m.}. The role of the threshold task, that implies the difference between the \emph{Tx} and the \emph{No Tx} bar for the \textit{CNN processing o.m.}, is more important. Even with this very high rate, the CNN-based monitoring is still convenient with respect to the \textit{raw data o.m.}, confirming the usefulness of near-sensor processing.

\subsection{Power model and operating mode power consumption estimation}
We have performed a thorough set of experiments measuring energy consumption in different setup conditions. The results were used to create a model highlighting the contribution of each task to the energy consumption of the node. By interpolating the experimental results on power consumption in the different use cases and knowing the duration of each task, we were able to build a model capable of estimating the energy consumption of the device under each possible use case. \hyperref[tab:taskCon]{Table \ref{tab:taskCon}} shows the energy values for each task in the process network. \hyperref[tab:platCon]{Table \ref{tab:platCon}} instead shows the power consumption of the platform in idle state and ECG sensor.
\begin{table}
	\centering
	\footnotesize
	\begin{tabular}{C{20mm} C{15mm} C{12mm} C{21mm}}
		\hline
		\\[-1.6mm]
		\textit{\textbf{Task type}} & \textit{\textbf{Number of cycle}} & \textit{\textbf{Execution time (\boldmath$8\,$MHz)}} & \textit{\textbf{Energy contribution}} \\
		\\[-3mm]
		\hline
		\hline
		\\[-2.5mm]
		\textit{\textbf{Get data}}      & $ 841 $       & $ 105 \,\mu s $   & $E_g = 2.96\,\mu J$   \\
		\\[-2.5mm]
		\textit{\textbf{Get data + peak}}          & $ 1\,550 + 841 $       & $ 300 \,\mu s $    & $ E_{gp} = 3.76\,\mu J$   \\
		\\[-2.5mm]
		\textit{\textbf{CNN 4\_4\_100}}     & $ 361\,360 $    & $ 45 \,ms $      & $ E_c = 148.78\,\mu J$ \\
		\\[-2.5mm]
		\textit{\textbf{CNN 20\_20\_100}}    & $ 1\,719\,582 $    & $  215\,ms $      & $ E_c = 660.37\,\mu J$ \\
		\\[-2.5mm]
		\textit{\textbf{Threshold}}     & $ 910 $       & $ 114 \,\mu s $   & $ E_t = 2.73\,\mu J$    \\
		\\[-2.5mm]
		\textit{\textbf{Send data}}     & $ \sim25\,000 $ & $ \sim3 \,ms $    & $ E_s = 83.96\,\mu J$  \\
        \\[-2.5mm]
		\hline
		\\[-1mm]
	\end{tabular}
	\caption{Summary of consumption and execution time for each tasks.}
	\label{tab:taskCon}
\end{table}
\begin{table}
	\centering
	\footnotesize
	\begin{tabular}{C{30mm} C{12mm} C{12mm} C{12mm}}
		\hline
		\\[-1.6mm]
		\textit{\textbf{Device}} & \multicolumn{3}{c}{\textit{\textbf{\shortstack{Power consumption}}}}\\
		\cmidrule(lr){2-4} &
		\boldmath$2\,M\!H\!z$ & \boldmath$4\,M\!H\!z$ & \boldmath$8\,M\!H\!z$ \\
		\\[-3mm]
		\hline
		\hline
		\\[-2.5mm]
		\textit{\textbf{Platform in idle state}}  & $ 2.609\,mW $ & $ 3.101\,mW $ & $ 4.546\,mW $ \\
		\\[-2.5mm]
		\textit{\textbf{ECG sensor}}              & $ 237\,uW $   & $ 237\,uW $   & $ 237\,uW $   \\
        \\[-2.5mm]
		\hline
		\\[-1mm]
	\end{tabular}
	\caption{Summary of consumption of peripherals.}
	\label{tab:platCon}
\end{table}

At this point it's possible to easily estimate the power consumption relative to each operating mode, the resulting equations that calculate the power consumption for each operating mode are:
\begin{itemize}
\item $P_{\textit{raw\,data\,o.m.}}$
\begin{equation*}
(E_g+ \alpha E_{s}) \cdot f_s + P_{idle} + P_{sensor}
\end{equation*}
\item $P_{\textit{peak\,detection\,o.m.}}$
\begin{equation*}
E_{gp} \cdot f_s + (E_t + E_s) \cdot f_{p} + P_{idle} + P_{sensor}
\end{equation*}
\item $P_{\textit{cnn\,processing\,o.m.}}$
\begin{equation*}
E_{gp} \cdot f_s + (E_c + E_t + E_s) \cdot f_{hr} + P_{idle} + P_{sensor}
\end{equation*}
\end{itemize}
where:
\begin{itemize}
\item $f_s$ is the sampling frequency,
\item $f_{hr}$ is the heart rate,
\item $f_{p}$ is the peak data sanding frequency,
\item $\alpha^{-1}$ it's the number of samples inserted in a BLE package,
\item $P_{idle}$ power consumption of the platform in idle state, depends on the system frequency,
\item $P_{sensor}$ energy consumption of the ECG sensor.
\end{itemize}

\begin{figure}
    \centering
    \includegraphics[width=0.58\textwidth]{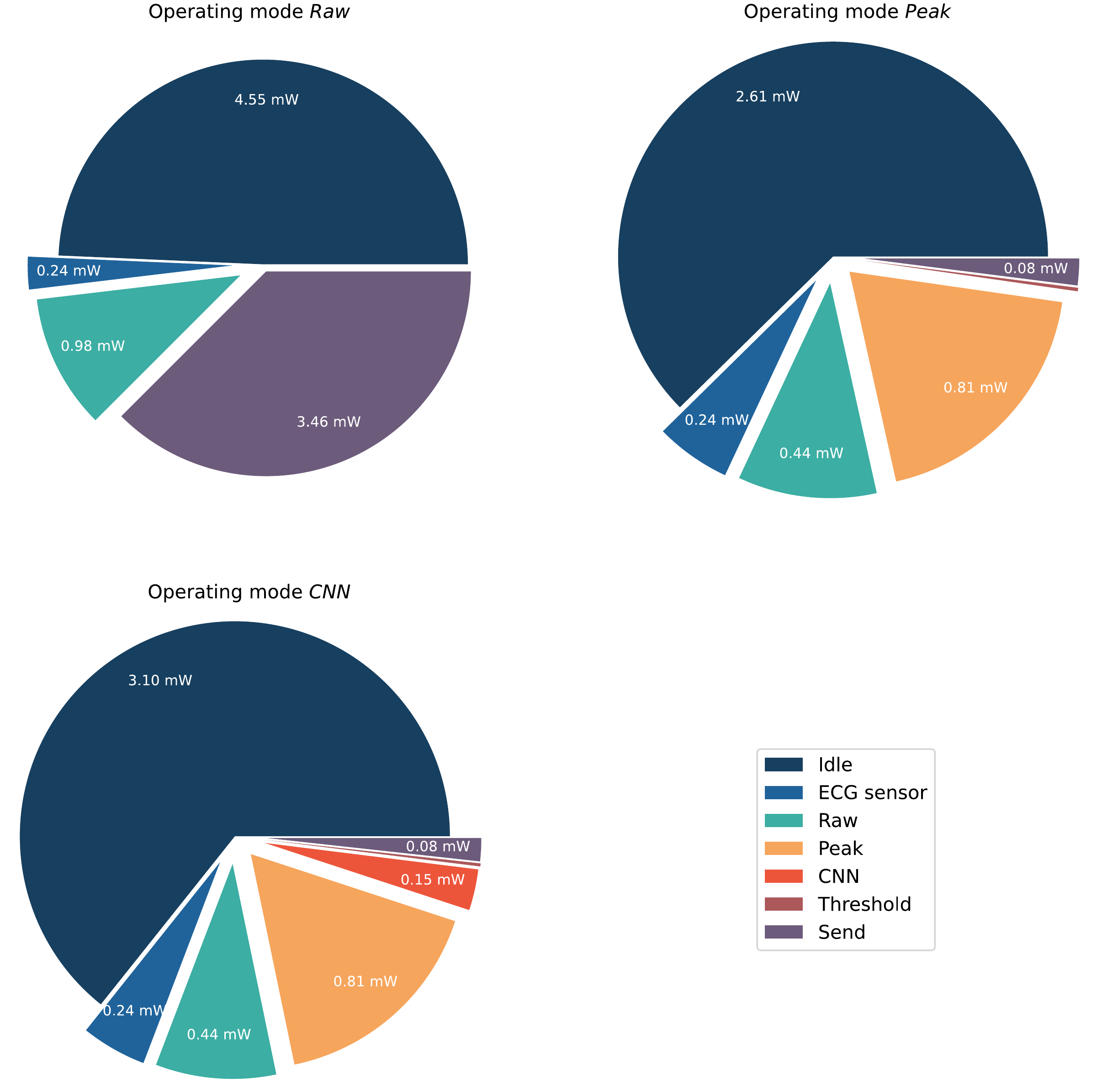}
    \caption{Estimation of energy consumption for each task of each operating modes at 60 bpm.}
    \label{fig:pie}
\end{figure}

\hyperref[fig:pie]{Figure \ref{fig:pie}} shows the estimate of the power consumption of the device and the contributions of each task in case the heart rate is around 60 $bpm$.
The following list shows the estimated battery life ($ 600 \, mAh $, $ 3.7 \, V $ Li-Ion) for each operating mode:
\begin{itemize}
\item \textit{Raw data o.m.}: 10.29 days
\item \textit{Peak detection o.m.}: 23.49 days
\item \textit{CNN processing o.m.}: 20.20 days
\end{itemize}

\section{Conclusion}
\label{sec:conclusion}
We have defined a hardware/software template for the development of a dynamically manageable IoMT node, studied to execute in-place analysis of the sensed physiological data. Its implementation has been tested on a low-power platform, able to exploit a CNN-based data analysis to recognize anomalies on ECG traces. The device is able to reconfigure itself according to the required operating modes and workload. The ADAM component, able to manage the reconfiguration of the device, plays a substantial role in energy saving. A quantized neural network reaches an accuracy value higher than 97\% on MIT-BIH Arrhythmia dataset for NLRAV and NSVFQ diseases classification. We measured an energy-saving up to 50\% by activating in-place analysis and managing the hardware and software components of the device. This work demonstrates how the feasibility of increasing battery lifetime with near-sensor processing and highlighting the importance of data-dependent runtime architecture management. 

\section*{Acknowledgment}
This work was supported by EU Commission for funding ALOHA
Project (H2020) under Grant Agreement n. 780788.
This work was also supported by the joint research and development project F/050395/01‐02/X32, INSIEME: INtelligent Systems for Integrated hEalth ManagEment, CUP:B28I17000060008, funded by Italian MISE (Ministero dello Sviluppo Economico), D.M. 01/06/2016, Axis 1, action 1.1.3. of the National Operative Program \guillemotleft Imprese e Competitività\guillemotright 2014‐2020 FESR, Horizon 2020 – PON I\&C 2014‐20.

\bibliographystyle{unsrt}  
\bibliography{ms}

\begin{thebibliography}{10}

\bibitem{cvd}
Emelia~J Benjamin, Paul Muntner, Alvaro Alonso, Marcio~S Bittencourt, Clifton~W
  Callaway, April~P Carson, Alanna~M Chamberlain, Alexander~R Chang, Susan
  Cheng, Sandeep~R Das, et~al.
\newblock Heart disease and stroke statistics—2019 update: a report from the
  american heart association.
\newblock {\em Circulation}, 139(10):e56--e528, 2019.

\bibitem{european}
Elizabeth Wilkins, L~Wilson, Kremlin Wickramasinghe, Prachi Bhatnagar, Jose
  Leal, Ramon Luengo-Fernandez, R~Burns, Mike Rayner, and Nick Townsend.
\newblock European cardiovascular disease statistics 2017.
\newblock 2017.

\bibitem{net1}
Zhe Yang, Qihao Zhou, Lei Lei, Kan Zheng, and Wei Xiang.
\newblock An iot-cloud based wearable ecg monitoring system for smart
  healthcare.
\newblock {\em Journal of Medical Systems}, 40(12):286, Oct 2016.

\bibitem{net3}
L.~{Roberts}, P.~{Michalák}, S.~{Heaps}, M.~{Trenell}, D.~{Wilkinson}, and
  P.~{Watson}.
\newblock Automating the placement of time series models for iot healthcare
  applications.
\newblock In {\em 2018 IEEE 14th International Conference on e-Science
  (e-Science)}, pages 290--291, Oct 2018.

\bibitem{Macis15}
Silvia Macis, Daniela Loi, Danilo Pani, Luigi Raffo, Serena~La Manna, Vincenzo
  Cestone, and Davide Guerri.
\newblock Home telemonitoring of vital signs through a tv-based application for
  elderly patients.
\newblock In {\em 2015 IEEE International Symposium on Medical Measurements and
  Applications (MeMeA) Proceedings}, pages 169--174, May 2015.

\bibitem{fit1}
Kanitthika Kaewkannate and Soochan Kim.
\newblock {\em The Comparison of Wearable Fitness Devices}.
\newblock 10 2018.

\bibitem{fit2}
Kanitthika Kaewkannate and Soo-Chan Kim.
\newblock A comparison of wearable fitness devices.
\newblock {\em BMC Public Health}, 16:1--16, 2016.

\bibitem{heavy1}
Ulas~Baran Baloglu, Muhammed Talo, Ozal Yildirim, Ru~San Tan, and U~Rajendra
  Acharya.
\newblock Classification of myocardial infarction with multi-lead ecg signals
  and deep cnn.
\newblock {\em Pattern Recognition Letters}, 122:23 -- 30, 2019.

\bibitem{heavy2}
Ruggero~Donida Labati, Enrique Muñoz, Vincenzo Piuri, Roberto Sassi, and Fabio
  Scotti.
\newblock Deep-ecg: Convolutional neural networks for ecg biometric
  recognition.
\newblock {\em Pattern Recognition Letters}, 2018.

\bibitem{heavy3}
Yazhao Li, Yanwei Pang, Jian Wang, and Xuelong Li.
\newblock Patient-specific ecg classification by deeper cnn from generic to
  dedicated.
\newblock {\em Neurocomputing}, 314:336 -- 346, 2018.

\bibitem{nodata}
Keith Marlon~R. Tabal, Felicito~S. Caluyo, and Joseph Bryan~G. Ibarra.
\newblock Microcontroller-implemented artificial neural network for
  electrooculography-based wearable drowsiness detection system.
\newblock In Hamzah~Asyrani Sulaiman, Mohd~Azlishah Othman, Mohd
  Fairuz~Iskandar Othman, Yahaya~Abd Rahim, and Naim~Che Pee, editors, {\em
  Advanced Computer and Communication Engineering Technology}, pages 461--472,
  Cham, 2016. Springer International Publishing.

\bibitem{bologna}
M.~{Magno}, M.~{Pritz}, P.~{Mayer}, and L.~{Benini}.
\newblock Deepemote: Towards multi-layer neural networks in a low power
  wearable multi-sensors bracelet.
\newblock In {\em 2017 7th IEEE International Workshop on Advances in Sensors
  and Interfaces (IWASI)}, pages 32--37, June 2017.

\bibitem{prev}
Matteo~Antonio Scrugli, Daniela Loi, Luigi Raffo, and Paolo Meloni.
\newblock A runtime-adaptive cognitive iot node for healthcare monitoring.
\newblock pages 350--357, 04 2019.

\bibitem{custom0}
A.~{Amirshahi} and M.~{Hashemi}.
\newblock Ecg classification algorithm based on stdp and r-stdp neural networks
  for real-time monitoring on ultra low-power personal wearable devices.
\newblock {\em IEEE Transactions on Biomedical Circuits and Systems},
  13(6):1483--1493, 2019.

\bibitem{custom1}
S.~{Lee}, J.~{Hong}, C.~{Hsieh}, M.~{Liang}, S.~{Chang Chien}, and K.~{Lin}.
\newblock Low-power wireless ecg acquisition and classification system for body
  sensor networks.
\newblock {\em IEEE Journal of Biomedical and Health Informatics},
  19(1):236--246, 2015.

\bibitem{custom2}
T.~{Chen}, E.~B. {Mazomenos}, K.~{Maharatna}, S.~{Dasmahapatra}, and
  M.~{Niranjan}.
\newblock Design of a low-power on-body ecg classifier for remote
  cardiovascular monitoring systems.
\newblock {\em IEEE Journal on Emerging and Selected Topics in Circuits and
  Systems}, 3(1):75--85, 2013.

\bibitem{custom3}
N.~{Bayasi}, T.~{Tekeste}, H.~{Saleh}, B.~{Mohammad}, A.~{Khandoker}, and
  M.~{Ismail}.
\newblock Low-power ecg-based processor for predicting ventricular arrhythmia.
\newblock {\em IEEE Transactions on Very Large Scale Integration (VLSI)
  Systems}, 24(5):1962--1974, 2016.

\bibitem{custom4}
Shuenn-Yuh Lee, Jia-Hua Hong, Cheng-Han Hsieh, Ming-Chun Liang, Shih-Yu~Chang
  Chien, and Kuang-Hao Lin.
\newblock Low-power wireless ecg acquisition and classification system for body
  sensor networks.
\newblock {\em IEEE Journal of Biomedical and Health Informatics},
  19(1):236--246, 2014.

\bibitem{custom5}
Turker Ince, Serkan Kiranyaz, and Moncef Gabbouj.
\newblock A generic and robust system for automated patient-specific
  classification of ecg signals.
\newblock {\em IEEE Transactions on Biomedical Engineering}, 56(5):1415--1426,
  2009.

\bibitem{custom6}
Eralp Kola{\u{g}}asio{\u{g}}lu.
\newblock Energy efficient feature extraction for single-lead ecg
  classification based on spiking neural networks.
\newblock 2018.

\bibitem{other1}
V.~{Natarajan} and A.~{Vyas}.
\newblock Power efficient compressive sensing for continuous monitoring of ecg
  and ppg in a wearable system.
\newblock In {\em 2016 IEEE 3rd World Forum on Internet of Things (WF-IoT)},
  pages 336--341, 2016.

\bibitem{other2}
U.~{Satija}, B.~{Ramkumar}, and M.~{Sabarimalai Manikandan}.
\newblock Real-time signal quality-aware ecg telemetry system for iot-based
  health care monitoring.
\newblock {\em IEEE Internet of Things Journal}, 4(3):815--823, 2017.

\bibitem{other3}
E.~{Spanò}, S.~{Di Pascoli}, and G.~{Iannaccone}.
\newblock Low-power wearable ecg monitoring system for multiple-patient remote
  monitoring.
\newblock {\em IEEE Sensors Journal}, 16(13):5452--5462, 2016.

\bibitem{other4}
G.~{Xu}.
\newblock Iot-assisted ecg monitoring framework with secure data transmission
  for health care applications.
\newblock {\em IEEE Access}, 8:74586--74594, 2020.

\bibitem{other5}
G.~R. {Deshmukh} and U.~M. {Chaskar}.
\newblock Iot enabled system design for real-time monitoring of ecg signals
  using tiva c-series microcontroller.
\newblock In {\em 2018 Second International Conference on Intelligent Computing
  and Control Systems (ICICCS)}, pages 976--979, 2018.

\bibitem{other6}
A.~{Walinjkar} and J.~{Woods}.
\newblock Personalized wearable systems for real-time ecg classification and
  healthcare interoperability: Real-time ecg classification and fhir
  interoperability.
\newblock In {\em 2017 Internet Technologies and Applications (ITA)}, pages
  9--14, 2017.

\bibitem{other7}
T.~{Shaown}, I.~{Hasan}, M.~M.~R. {Mim}, and M.~S. {Hossain}.
\newblock Iot-based portable ecg monitoring system for smart healthcare.
\newblock In {\em 2019 1st International Conference on Advances in Science,
  Engineering and Robotics Technology (ICASERT)}, pages 1--5, 2019.

\bibitem{other8}
U.~{Arun}, S.~{Natarajan}, and R.~R. {Rajanna}.
\newblock A novel iot cloud-based real-time cardiac monitoring approach using
  ni myrio-1900 for telemedicine applications.
\newblock In {\em 2018 3rd International Conference on Circuits, Control,
  Communication and Computing (I4C)}, pages 1--4, 2018.

\bibitem{lp1}
Hassan Ghasemzadeh and Roozbeh Jafari.
\newblock Ultra low-power signal processing in wearable monitoring systems: A
  tiered screening architecture with optimal bit resolution.
\newblock {\em ACM Trans. Embed. Comput. Syst.}, 13(1):9:1--9:23, September
  2013.

\bibitem{lp2}
T.~{Tekeste}, H.~{Saleh}, B.~{Mohammad}, and M.~{Ismail}.
\newblock Ultra-low power qrs detection and ecg compression architecture for
  iot healthcare devices.
\newblock {\em IEEE Transactions on Circuits and Systems I: Regular Papers},
  66(2):669--679, Feb 2019.

\bibitem{lp3}
C.~{Wang}, Y.~{Qin}, H.~{Jin}, I.~{Kim}, J.~D. {Granados Vergara}, C.~{Dong},
  Y.~{Jiang}, Q.~{Zhou}, J.~{Li}, Z.~{He}, Z.~{Zou}, L.~{Zheng}, X.~{Wu}, and
  Y.~{Wang}.
\newblock A low power cardiovascular healthcare system with cross-layer
  optimization from sensing patch to cloud platform.
\newblock {\em IEEE Transactions on Biomedical Circuits and Systems}, pages
  1--1, 2019.

\bibitem{lp4}
Mahesh~Kumar Adimulam and M.~B. Srinivas.
\newblock Ultra low power programmable wireless exg soc design for iot
  healthcare system.
\newblock In Paolo Perego, Amir~M. Rahmani, and Nima TaheriNejad, editors, {\em
  Wireless Mobile Communication and Healthcare}, pages 41--49, Cham, 2018.
  Springer International Publishing.

\bibitem{ieee1}
M.~{Deshmane} and S.~{Madhe}.
\newblock Ecg based biometric human identification using convolutional neural
  network in smart health applications.
\newblock In {\em 2018 Fourth International Conference on Computing
  Communication Control and Automation (ICCUBEA)}, pages 1--6, 2018.

\bibitem{ieee2}
K.~G. {Rani Roopha Devi}, R.~{Mahendra Chozhan}, and R.~{Murugesan}.
\newblock Cognitive iot integration for smart healthcare: Case study for heart
  disease detection and monitoring.
\newblock In {\em 2019 International Conference on Recent Advances in
  Energy-efficient Computing and Communication (ICRAECC)}, pages 1--6, 2019.

\bibitem{ieee6}
S.~{Sakib}, M.~M. {Fouda}, Z.~M. {Fadlullah}, and N.~{Nasser}.
\newblock Migrating intelligence from cloud to ultra-edge smart iot sensor
  based on deep learning: An arrhythmia monitoring use-case.
\newblock In {\em 2020 International Wireless Communications and Mobile
  Computing (IWCMC)}, pages 595--600, 2020.

\bibitem{acc}
Y.~{Xitong}, D.~{Yu}, and Z.~{Jianxun}.
\newblock A real - time ecg signal classification algorithm.
\newblock In {\em 2020 39th Chinese Control Conference (CCC)}, pages
  7356--7361, 2020.

\bibitem{ieee5}
I.~{Azimi}, J.~{Takalo-Mattila}, A.~{Anzanpour}, A.~M. {Rahmani},
  J.~{Soininen}, and P.~{Liljeberg}.
\newblock Empowering healthcare iot systems with hierarchical edge-based deep
  learning.
\newblock In {\em 2018 IEEE/ACM International Conference on Connected Health:
  Applications, Systems and Engineering Technologies (CHASE)}, pages 63--68,
  2018.

\bibitem{ieee3}
A.~{Burger}, C.~{Qian}, G.~{Schiele}, and D.~{Helms}.
\newblock An embedded cnn implementation for on-device ecg analysis.
\newblock In {\em 2020 IEEE International Conference on Pervasive Computing and
  Communications Workshops (PerCom Workshops)}, pages 1--6, 2020.

\bibitem{rbm}
Sherin~M. Mathews, Chandra Kambhamettu, and Kenneth~E. Barner.
\newblock A novel application of deep learning for single-lead ecg
  classification.
\newblock {\em Computers in Biology and Medicine}, 99:53--62, 2018.

\bibitem{san}
G.~Sannino and G.~{De Pietro}.
\newblock A deep learning approach for ecg-based heartbeat classification for
  arrhythmia detection.
\newblock {\em Future Generation Computer Systems}, 86:446--455, 2018.

\bibitem{dcnn}
Serkan Kiranyaz, Turker Ince, and Moncef Gabbouj.
\newblock Real-time patient-specific ecg classification by 1-d convolutional
  neural networks.
\newblock {\em IEEE Transactions on Biomedical Engineering}, 63(3):664--675,
  2016.

\bibitem{rasp}
Dennis Hou, M.D. Raymond~Hou, and Janpu Hou.
\newblock Ecg beat classification on edge device.
\newblock In {\em 2020 IEEE International Conference on Consumer Electronics
  (ICCE)}, pages 1--4, 2020.

\bibitem{Lai2018}
Liangzhen Lai, Naveen Suda, and Vikas Chandra.
\newblock {CMSIS-NN:} efficient neural network kernels for arm cortex-m cpus.
\newblock {\em CoRR}, abs/1801.06601, 2018.

\bibitem{cnn}
D.~{Li}, J.~{Zhang}, Q.~{Zhang}, and X.~{Wei}.
\newblock Classification of ecg signals based on 1d convolution neural network.
\newblock In {\em 2017 IEEE 19th International Conference on e-Health
  Networking, Applications and Services (Healthcom)}, pages 1--6, Oct 2017.

\bibitem{pt}
J.~{Pan} and W.~J. {Tompkins}.
\newblock A real-time qrs detection algorithm.
\newblock {\em IEEE Transactions on Biomedical Engineering},
  BME-32(3):230--236, 1985.

\end{thebibliography}

\end{document}